\documentclass[letterpaper]{article} 
\usepackage[draft]{aaai25}  
\usepackage{times}  
\usepackage{helvet}  
\usepackage{courier}  
\usepackage[hyphens]{url}  
\usepackage{graphicx} 
\usepackage{natbib}  
\usepackage{caption} 
\usepackage{algorithm}
\usepackage{algorithmic}

\usepackage{newfloat}
\usepackage{xcolor,colortbl}
\usepackage{booktabs}       
\usepackage{listings}
\usepackage{amsmath, amssymb}
\usepackage{xspace}
\usepackage{cleveref}
\usepackage{tabularx}
\usepackage{nicefrac}       
\usepackage{microtype}      
\usepackage{multirow}
\usepackage{subcaption}
\usepackage{pifont}
\usepackage{textcomp}
\usepackage{array}
\newcolumntype{P}[1]{>{\centering\arraybackslash}p{#1}}
\newcommand{\cmark}{\ding{51}}%
\newcommand{\xmark}{\ding{55}}%

\definecolor{lbcolor}{rgb}{0.95,0.95,0.95}
\usepackage[nolist,nohyperlinks]{acronym}

\DeclareCaptionStyle{ruled}{labelfont=normalfont,labelsep=colon,strut=off} 
\floatstyle{ruled}
\newfloat{listing}{tb}{lst}{}
\floatname{listing}{Listing}
\title{CriSPO: Multi-Aspect Critique-Suggestion-guided \\Automatic Prompt Optimization for Text Generation\\ {\small(Extended version)}}

\author{
Han He\equalcontrib, Qianchu Liu\equalcontrib, Lei Xu\equalcontrib, Chaitanya Shivade, \\Yi Zhang, Sundararajan Srinivasan, Katrin Kirchhoff\\
}
\affiliations{
    Amazon AWS AI Labs\\
    \{hankcs, liufqian, leixx, shivadc, \\yizhngn, sundarsr, katrinki\}@amazon.com
}

\newcommand*{\ours}{CriSPO\xspace}
\newcommand*{\fcrispo}{\mathcal{F}}
\newcommand*{\bestprompt}{p^*}
\newcommand*{\ul}{\underline}

\begin{document}
\maketitle

\acrodef{LLM}{Large Language Model}
\acrodef{OPRO}{Optimization by PROmpting}
\acrodef{AST}{Automatic Suffix Tuning}
\acrodef{CoT}{Chain-of-Thought}
\acrodef{QA}{Question Answering}
\acrodef{RAG}{Retrieval-Augmented Generation}
\acrodef{ICL}{In-Context Learning}

\begin{abstract}
Existing automatic prompt engineering methods are typically designed for discriminative tasks, where new task prompts are iteratively refined with limited feedback from a single metric reflecting a single aspect. 
However, these approaches are suboptimal for generative tasks, which require more nuanced guidance beyond a single numeric metric to improve the prompt and optimize multiple aspects of the generated text.
To address these challenges, we propose a novel multi-aspect \textbf{Cri}tique-\textbf{S}uggestion-guided automatic \textbf{P}rompt \textbf{O}ptimization (\ours) approach. \ours introduces a \textit{critique-suggestion module} as its core component. This module spontaneously discovers aspects, and compares generated and reference texts across these aspects, providing specific suggestions for prompt modification. These clear critiques and actionable suggestions guide a \textit{receptive optimizer module} to make more substantial changes, exploring a broader and more effective search space. 
To further improve \ours with multi-metric optimization, we introduce an Automatic Suffix Tuning (AST) extension to enhance the performance of task prompts across multiple metrics.
We evaluate \ours on 4 state-of-the-art \acp{LLM} across 4 summarization and 5 \ac{QA} datasets. Extensive experiments show 3-4\% ROUGE score improvement on summarization and substantial improvement of various metrics on \ac{QA}.
\end{abstract}

\begin{links}
\link{Code}{https://github.com/amazon-science/CriSPO}
\end{links}

\section{Introduction}
\acp{LLM} have emerged as powerful tools for various natural language processing tasks, including text generation \cite{brown2020language}. To fully leverage their capabilities, a critical step is to design a precise \textit{task prompt} which specifies the desired behavior of the \ac{LLM} to solve a task. Manual prompt engineering is often laborious, skill-intensive and sub-optimal, motivating the need for automatic prompt engineering techniques which automatically tune the task prompt.

Recent research has made notable progress in automatic prompt engineering for discriminative tasks, such as text classification \cite{zhou2022large,yang2023large,pryzant-etal-2023-automatic,sordoni2024joint}. These methods focus on optimizing task prompts for a single metric on a single aspect. The process typically involves instructing an \ac{LLM} optimizer with a meta-prompt to generate new task prompts based on previously sampled task prompts and their corresponding scores. By iteratively exploring candidates and selecting the task prompt with the highest score, performance on the target metric improves over numerous iterations. However, applying these methods directly to text generation tasks, such as summarization, is sub-optimal due to challenges in obtaining effective \textit{optimization signals}. Unlike classification tasks, where metrics are straightforward (eg. accuracy), automatic metrics for text generation, like ROUGE~\cite{lin-2004-rouge},
provides limited guidance for prompt refinement. 
For example, a lower ROUGE score may result from aspects such as mismatched length, differences in word choice due to formality, or varying writing formats, making it difficult to guide LLMs in prompt modification without fine-grained feedback targeting these individual aspects. 
Furthermore, evaluating text generation involves multiple metrics \cite{fabbri-etal-2021-summeval, gao-wan-2022-dialsummeval, elangovan2024considers}. In addition to reference similarity, other metrics such as factual consistency, which can be assessed using metrics like AlignScore~\cite{zha-etal-2023-alignscore}, are also important. Balancing or utilizing these multiple metrics is not fully addressed by existing prompt engineering methods that focus on optimizing a single metric.

To address these challenges, we introduce \ours, a multi-aspect \textbf{Cri}tique-\textbf{S}uggestion-guided automatic \textbf{P}rompt \textbf{O}ptimization (\ours) approach. Overall, our approach employs \acp{LLM} to automatically identifies multi-aspect prompt revision suggestions, based on which prompts are automatically designed and refined (\Cref{tab:example} in Appendix shows a working example of how a prompt gets revised in \ours). Inspired by recent self-reflection studies, where LLMs generate verbal feedback to aid in self-improvement \cite{gero2023self,NEURIPS2023_1b44b878,madaan2024self}, we designed the first key component of \ours: the \textbf{multi-aspect critique-suggestion meta-prompt}. It automatically discovers proper aspects to compare generated text with reference, write critiques of flaws \cite{pryzant-etal-2023-automatic} and suggestions to improve the task prompt (\Cref{fig:word_cloud} shows a word cloud of aspects identified by \ours, including number of words, style, and precision). Both critiques and suggestions, written in natural language, are more helpful for prompt improvement than a single ROUGE score. We then create a \textbf{receptive optimizer meta-prompt} that generates new prompts. In addition to conditioning on previous high-score task prompts and scores, this optimizer also reviews the past critiques and suggestions. It then generates an overall suggestion and an improved task prompt candidate in a \ac{CoT} \cite{wei2022chain} manner. Our approach iteratively optimizes the task prompt using \acp{LLM} similar to previous work like \ac{OPRO}~\cite{yang2023large}, but it enriches the training signal with multi-aspect critiques and suggestions to better optimize a text generation metric. 
To further enhance performance by allowing the prompt to access external data, we design the \textbf{task prompt template} that contains placeholders for \ac{ICL} examples or retrieved contexts. The receptive optimizer meta-prompt generates these templates directly, so it can flexibly move components in task prompt for better organization.

While \ours offers multi-aspect guidance for optimizing text generation through critiques and suggestions, we further enhance this guidance by incorporating multiple metrics as additional teaching signals. To this end, we propose a novel \ac{AST} extension which divides prompts into chunks conquering different metrics. 
Through multi-objective learning, we improve each new metric with little to no drop in existing metrics.

We test \ours on state-of-the-art \acp{LLM}, including Claude \citep{claude_instant_model, anthropic2024claude}, Mistral \citep{jiang2023mistral} and Llama3 \cite{llama3_model}, across 9 heterogeneous datasets. These include 4 summarization datasets spanning various abstractiveness, formats, and domains, as well as 5 \ac{QA} datasets. Extensive experiments demonstrate that \ours significantly improves prompt quality and task performance over strong baselines as verified by human evaluation. We also conduct ablation study to assess the effectiveness of key ingredients.

\noindent \textbf{Our contributions are summarized below: }

\textbf{1)} We propose \ours, an automatic prompt engineering approach tailored for generative tasks. It discovers aspects to critique generated text and write suggestions for more effective prompt revision.

\textbf{2)} We conduct comprehensive experiments across multiple \acp{LLM} and datasets, demonstrating the effectiveness and robustness of our method.
We show an overall 3-4\% improvement on ROUGE scores with qualitative verification from human evaluation. \ours also obtained consistent improvements on various QA tasks. 

\textbf{3)} We propose \ac{AST} that enables prompt tuning for multiple metrics. We show that \ours  with \ac{AST} can jointly optimize AlignScore~\cite{zha-etal-2023-alignscore} for faithfulness and ROUGE for reference similarity.



\section{Related Work}
\begin{figure*}[t]
    \centering
    \includegraphics[width=.8\textwidth]{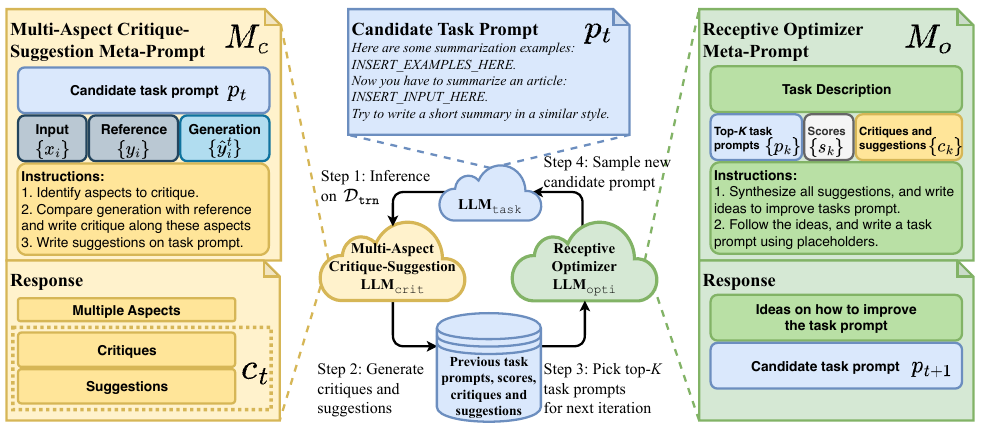}
    \caption{The \ours workflow for text generation tasks. In each iteration, a candidate task prompt $p_t$ is applied to $\mathcal{D}_{\texttt{trn}}$ (step~1) and evaluated using a multi-aspect critique-suggestion meta-prompt $M_c$ (step~2). We select top-$K$ previously sampled task prompts (step~3) and use a receptive optimizer $M_o$ to generate the next candidate $p_{t+1}$ (step~4). The automatic optimization loop runs multiple iterations, while the best task prompt is selected based on performance on $\mathcal{D}_{\texttt{dev}}$.}
    \label{fig:framework}
\end{figure*}
There is an increasing effort in the literature to explore gradient-free automatic prompt engineering methods with off-the-shelf \acp{LLM}. The focus of these approaches is to find a good search algorithm for better prompt candidates to solve discriminitive tasks. Earlier studies have employed conventional paraphrasing methods for prompt generation through editing phrases \citep{prasad-etal-2023-grips} or back translation \citep{xu-etal-2022-gps}. More recently, \acp{LLM} themselves have been used to sample prompt candidates. \citet{zhou2022large} proposed Automatic Prompt Engineering (APE) which iteratively prompts an \ac{LLM} to generate semantically similar variations of the locally best prompt. \citet{pryzant-etal-2023-automatic} add verbal feedback based on error examples to propose better prompts in terms of accuracy. Concurrently, \citet{sordoni2024joint} learn prompts with variational
inference by considering their outputs as latent variables. Later on, \citet{yang2023large} propose \ac{OPRO} to improve over them by incorporating the history of past prompts with their scores which stabilizes optimization. More structured prompts have also been explored by imposing expert-level planning \citep{wang2023promptagent}. In a parallel thread, \citet{fernando2023promptbreeder} and \citet{guo2023connecting} were inspired by evolutionary algorithms to perform mutation operations for prompt generation. All of the existing approaches have mostly been designed to target classification tasks using a single metric.
 Comparing to the existing studies, our proposed method specifically targets the unique challenges in text generation and approaches the prompt optimization problem in a multi-aspect and multi-metric fashion. 
For practitioners, \citet{khattab2023dspy} design DSPy framework to build and optimize complex \ac{LLM} pipelines in a programmatic fashion. TextGrad \cite{yuksekgonul2024textgrad} further generalizes  optimization to text beyond prompt. Our \ours can be used as a powerful optimizer in these frameworks.

Our approach is also inspired by recent studies on using \ac{LLM}s to automatically correct its output \citep{pan2023automatically, madaan2024self}. \citet{gero2023self} apply multiple self-reflection steps to improve the performance of information extraction.  \citet{yan-etal-2024-predicting} use \ac{CoT} to generate structured comparison and preferences for two model outputs. \citet{NEURIPS2023_1b44b878} argue the importance of the self-reflection history and propose reflexion agent to provide verbal feedback on past trials for better decision in the next trials. It is important to notice that these self-reflection studies are strictly speaking not automatic prompt engineering approaches as these studies optimize output revision rather than directly on the prompts.
\ours, however, automatically reflects on the design of the prompt and uses these past reflections to revise the prompts.




\section{Method}
\label{sec:method}

\noindent\textbf{Problem Formulation:} In a text generation task, let \(\mathcal{D}_{\texttt{trn}} = \{(x_i, y_i)\}_{i=1\ldots n}\) be the training set, with a development set \(\mathcal{D}_{\texttt{dev}}\) and a test set \(\mathcal{D}_{\texttt{tst}}\). Here, \(x\) represents the input data, and \(y\) is the corresponding ground truth reference. A \textbf{task prompt} \(p\) comprises instructions that, when filled with input \(x\), are fed to a black-box API \(\text{LLM}_{\text{task}}\)\footnote{We use notations \(\text{LLM}_{\text{task}}, \text{LLM}_{\text{crit}}, \text{LLM}_{\text{opti}}\) for clarity. Though they share the same underlying \ac{LLM} unless specified otherwise.} to generate a completion \(\hat{y} = \text{LLM}_{\text{task}}(p, x)\). The goal is to optimize \(p\) using \(\mathcal{D}_{\texttt{trn}}\) and \(\mathcal{D}_{\texttt{dev}}\) to identify an optimal prompt \(\bestprompt\) that maximizes performance on one or more evaluation metrics on \(\mathcal{D}_{\texttt{tst}}\).

\noindent\textbf{\ours Overview:} \ours is an automatic prompt optimization algorithm designed to iteratively re{\underline f}ine a task prompt \(p\) from an initial seed prompt \(p_0\) to the optimum: $\bestprompt\leftarrow \fcrispo(p_0)$. In each iteration $t$, we conduct the following steps:
\begin{itemize}
    \item \ul{Evaluate on \(\mathcal{D}_\texttt{trn}\)}: Apply the candidate prompt \(p_t\) on \(\mathcal{D}_\texttt{trn}\), call \(\text{LLM}_{\texttt{task}}\) to generate outputs \(\{\hat{y}^t_i\}_{i=1\ldots n}\) and compute a primary metric \(s_t\), which can be a single metric or an aggregation of multiple metrics.
    \item \ul{Generate Critiques and Suggestions:} Apply the \textbf{multi-aspect critique-suggestion} meta-prompt \(M_c\)  and call  \(\text{LLM}_{\texttt{crit}}\) to compare  \(\{\hat{y}^t_i\}_{i=1\ldots n}\)  and  \(\{{y}^t_i\}_{i=1\ldots n}\) and generate critiques and suggestions \(c_t\) (see Section~\ref{sec:critique}).
    \item \ul{Generate a Candidate Task Prompt:} Select the top-\(K\) task prompts from previous iterations based on the primary metric, and insert the corresponding \(K\) triples \(\{(p_k, s_k, c_k)\}\) into the \textbf{receptive optimizer} meta-prompt \(M_o\). Then call \(\text{LLM}_{\texttt{opti}}\) to generate the next candidate prompt \(p_{t+1}\) (see Section~\ref{sec:prompt-optimization}).

\end{itemize}
We evaluate the current prompt \(p_t\) on \(\mathcal{D}_{\texttt{dev}}\) and select \(\bestprompt\) based on the primary metric. Upon reaching the maximum number of iterations, we apply an optional \ac{AST} to enhance performance on secondary metrics on  \(\bestprompt\) (see Section~\ref{sec:multi-metric}).
Figure~\ref{fig:framework} demonstrates the workflow of \ours on summarization tasks. Table~\ref{tab:example} (in Appendix) shows a concrete working example of \ours. 

\subsection{Multi-Aspect Critiques and Suggestions}
\label{sec:critique}
Given a prompt $p_t$ and its outputs $\{\hat{y}^t_i\}$ on \(\mathcal{D}_{\texttt{trn}}\), we design a multi-aspect critique-suggestion meta-prompt $M_c$ to identify \textit{critiques} -- flaws of the generated outputs across multiple aspects, and \textit{suggestions} -- specific edits on the task prompt to rectify each flaw.

\begin{figure}[htb]
\centering\includegraphics[width=0.9\columnwidth]{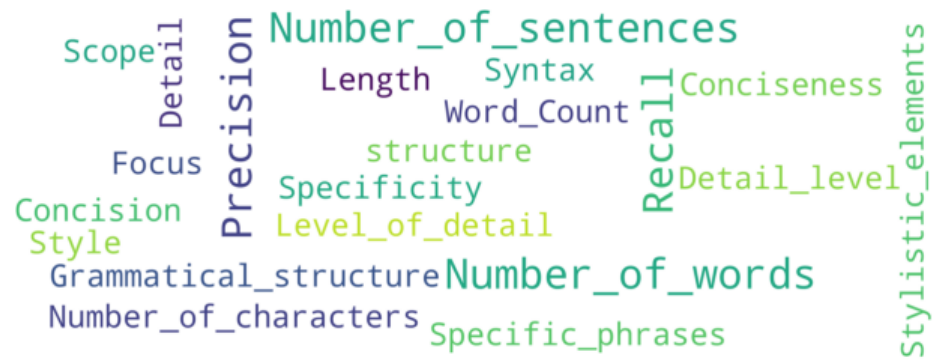}
        \caption{A word cloud showing the different aspects identified by \ours when comparing generations and references. }
        \label{fig:word_cloud}
\end{figure}

\paragraph{Constructive critiques with spontaneous dimension discovery:} In $M_c$, we first instruct $\text{LLM}_{\texttt{crit}}$ to generate several task-specific and iteration-specific aspects for a given batch of outputs from the current $p_t$. This approach ensures that as task prompts evolve across iterations, the focus remains on relevant aspects, addressing specific issues that arise. \Cref{fig:word_cloud} illustrates the aspects discovered during optimization. For each aspect, $M_c$ instructs $\text{LLM}_{\texttt{crit}}$ to generates a critique highlighting potential problems of the outputs generated with $p_t$ on the batch. 

\paragraph{Multi-aspect suggestions:} In line with each critique, a corresponding suggestion is made by $\text{LLM}_{\texttt{crit}}$ to edit $p_t$. As opposed to \citet{pryzant-etal-2023-automatic}, we decoupled the edit suggestion module from the new prompt generation process. Rather than generating a new prompt with each suggestion, we pack a history of critiques and suggestions into the receptive optimizer for generating the next prompt, enabling more stable optimization over the infinite search space.

Our $M_c$ is implemented in a single \ac{CoT} meta-prompt which generates, dimensions, critiques and suggestions in one single LLM call, specifically
\[ c_t=\text{LLM}_{\text{crit}}\left(M_c, p_t, (x_i, y_i, \hat{y}^t_i)_{i=1\ldots n}\right).\]
$M_c$ for different \acp{LLM} and tasks are shown in \Cref{sec:critique-prompt}.

\begin{table*}[tb]
  \centering\small
\small
  \begin{tabular}{@{}m{1.5cm}m{1.3cm}m{0.4cm}m{0.4cm}m{0.2cm}m{0.01cm}m{0.4cm}m{0.4cm}m{0.2cm}m{0.01cm}m{0.4cm}m{0.4cm}m{0.2cm}m{0.01cm}m{0.4cm}m{0.4cm}m{0.2cm}m{0.01cm}m{0.4cm}m{0.2cm}c@{}}
    \toprule
    & & \multicolumn{6}{c}{\textbf{Manual}} & \multicolumn{13}{c}{\textbf{Automatic Prompt Engineering}} \\
    \cmidrule(lr){3-10} \cmidrule(lr){11-21}

     && \multicolumn{3}{c}{\textbf{0-shot}} && \multicolumn{3}{c}{\textbf{3-shot*}}  &&\multicolumn{3}{c}{\textbf{OPRO}} && \multicolumn{3}{c}{\textbf{\ours}} &&\multicolumn{3}{c}{\textbf{\ours 3-shot*}} \\
     \cmidrule(lr){3-6} \cmidrule(lr){7-10} 
     \cmidrule(lr){11-14}
     \cmidrule(lr){15-18}
          \cmidrule(lr){19-21}

      \textbf{Dataset} & \textbf{\ac{LLM}} & \textbf{R1} & \textbf{R2} & \textbf{RL} && \textbf{R1} & \textbf{R2} & \textbf{RL}  &&\textbf{R1} & \textbf{R2} & \textbf{RL} && \textbf{R1} & \textbf{R2} & \textbf{RL} &&\textbf{R1} & \textbf{R2} & \textbf{RL} \\
    \midrule
    CNN & Claude In. & 37.5 & 12.5 & 22.6 && 40.4 & 14.8& 24.8 && 39.5 & 14.3 & 24.5 && 40.1 & 15.7 & 26.1 &&  {\bf 42.1} & {\bf 17.0} &{\bf 27.4}\\
    & Claude3 & 38.8 &14.4 &24.0 && 40.3&15.4&25.2 && 39.7&15.1&5.1&&  {\bf 42.2}&{\bf17.3}&{\bf27.9} && 41.6 &16.3 &27.1\\
    & Mistral {\scriptsize 7B} & 30.9 &11.0&20.4 && 30.7&10.6&20.1 && 36.5&{\bf14.4}&23.0 && 38.5&14.3&23.9 && {\bf 38.5}&14.3&{\bf24.1}\\
    & Llama3 {\scriptsize 8B} & 37.9 & 14.4 & 23.8 &&  &  & &&39.1&15.2&24.6\textsuperscript{\#} &&{\bf41.5}&{\bf 16.3}&{\bf 26.5}\textsuperscript{\#} &&&&\\
    \addlinespace
    MeetingBank & Claude In. & 30.7&11.6&20.5 && 34.2&17.3&25.5 &&39.0&20.3&29.7 && 41.4&23.7&33.1 && {\bf 50.1}&{\bf 35.4}&{\bf 44.4}  \\
    & Claude3 & 31.2&14.2&22.3 &&37.5&22.0&29.5 && 41.5&21.8&32.0 && 47.4&32.5&40.9 && {\bf 58.5}&{\bf 46.5}&{\bf 54.1} \\
    & Mistral {\scriptsize 7B} & 26.0&11.5&18.5 && 31.3&14.8&22.7 && 33.9&15.4&24.2  && {\bf 39.1}&{\bf 19.5}&{\bf 29.3} && 35.2& 16.7&26.1  \\
    & Llama3 {\scriptsize 8B}  & 31.4&14.6&22.6&&&&&&  40.2&22.3&31.5\textsuperscript{\#} && {\bf 44.7}&{\bf27.6}&{\bf36.8}\textsuperscript{\#} &&&&\\
    \addlinespace
    {SAMSum} & Claude In. & 33.9&11.7&25.6&& 37.8&14.3&28.8 & &38.1&13.4&28.7 & &44.4&16.9&34.3 && {\bf 45.7} &{\bf18.7} &{\bf 36.2}  \\
    & Claude3 & 35.8&12.7&27.0 && 41.1&16.6&31.3 && 39.0&14.7&30.1 && 43.4 &17.1&34.3&& {\bf 47.2} &{\bf 20.8} &{\bf38.2} \\
    & Mistral {\scriptsize 7B} & 32.0&10.2&24.1&& 39.5&14.1&30.3 & &37.9&13.6&29.0 & &37.6&12.4&28.4 & &{\bf40.0}&{\bf 14.2}&{\bf 30.8} \\
    & Llama3 {\scriptsize 8B} & 35.7&12.3&27.1& &&&&&39.3&14.7&30.0\textsuperscript{\#} && {\bf 44.8}&{\bf 18.8} &{\bf 35.4}\textsuperscript{\#} &&&&\\
    \addlinespace
    ACI-Bench & Claude In. & 43.8&16.9&26.1& &51.5&23.6&33.5 && 45.2&16.3&25.5 & &53.0&19.7&26.8 && {\bf 58.2} &{\bf 26.7}&{\bf35.3} \\
    & Claude3 & 47.3&20.3&29.3 && 59.1&30.1&38.6 && 48.8&20.1&29.5&& 54.0&21.4&30.3&&  {\bf 63.1}&{\bf 32.5}&{\bf41.0}\\
    & Mistral {\scriptsize 7B} & 47.8&17.7&25.4 && 48.4&{\bf 19.2}&{\bf28.1} &&45.1&17.0&25.2  && 50.2&18.2&25.6   && {\bf 50.3}&18.7&26.2\\
    & Llama3 {\scriptsize 8B} & 50.5&19.8&27.7 &&&&& & 54.2&22.0&29.3\textsuperscript{\#} && {\bf 56.2}&{\bf22.8}&{\bf29.9}\textsuperscript{\#} &&&&\\
    \midrule
     Average & Claude In. & 36.5 &   13.2 &   23.7&&41.0   & 17.5  &  28.2&&40.4 &   16.1  &  27.1&&44.7 &    19.0 &   30.1 &&{\bf 49.0}  &  {\bf 24.4} &   {\bf 35.8}\\
    & Claude3 & 38.3 &   15.4 &   25.6&&
    44.5  &  21.0  &  31.2&&
    42.2  &  17.9   & 29.2&&46.8   & 22.1  &  33.3&&{\bf 52.6} &   {\bf 29.0}  &  {\bf 40.1}\\
    & Mistral {\scriptsize 7B} &34.2 &    12.6  &  22.1 &&
    37.5 &   14.7   & 25.3&&
    38.4  &  15.1   & 25.4&&
    {\bf 41.4}   & {\bf 16.1}  &  {\bf 26.8}&&
    41.0  &  16.0 &  {\bf 26.8}\\
    & Llama3 {\scriptsize 8B} & 38.9 &   15.3 &   25.3&&&&&&
    43.2 &   18.6    &28.8&&
    {\bf 46.8}  &  {\bf 21.4}   & {\bf 32.2}&&&&
    \\
    \bottomrule
  \end{tabular}%
  \caption{Comparing \ours with manual prompts and \ac{OPRO} on representative summarization benchmarks. Averaged R1/R2/RL (i.e. ROUGE-1/2/L) are reported  across 3 runs. 3-shot*: 3-shot \ac{ICL} with example selection. Claude3: Claude3 Sonnet. Claude In.: Claude Instant. Llama3 results marked with (\#) are using Claude3 Sonnet as the optimizer. 3-shot* results are empty for Llama3 due to its limited context window. Standard deviation and SOTA results are in Appendix~\ref{appendix:full metrics}. Claude3 Sonnet achieves new SOTA ROUGE-1 performance on ACI-Bench. 
  }
  \label{tab:main_result}
\end{table*}

\subsection{Receptive Prompt Optimizer}
\label{sec:prompt-optimization}
Our receptive prompt optimizer meta-prompt $M_o$ improves over the OPRO optimizer meta-prompt \citep{yang2023large} by enriching its optimization trajectory $\{(p_k, s_k)\}$ with past critiques and suggestions $c_k$. Thus, ours samples candidate prompts for the next iteration conditioned on an enriched optimization trajectory: 
\[p_{t+1}=\text{LLM}_{\text{opti}}\left(M_o, \{(p_k, s_k, c_k)\}\right).\]
Specifically, we enhance the OPRO optimizer module with the following three improvements to better utilize critiques and suggestions for achieving stronger guidance and better exploration. See \Cref{sec:meta-prompt} for all $M_o$ by \acp{LLM} and tasks.


\paragraph{Enriched optimization trajectory:} The critiques and suggestions generated in Section~\ref{sec:critique} are used in an enriched optimization trajectory to propose new prompts via an OPRO-style optimizer. Specifically, our enriched optimization trajectory includes the top-$K$ best-performing past prompts $\{p_k\}$, their scores $\{s_k\}$, critiques and suggestions $\{c_k\}$, sorted in the ascending order by scores. Including critiques and suggestions in the optimization trajectory allows the LLM to avoid common limitations and identify common strengths from the past prompts for stable optimization.

\paragraph{Chain-of-thought:} After enriching the optimization trajectory, we also apply \ac{CoT} to the optimization process. Specifically, $\text{LLM}_{\texttt{opti}}$ is explicitly asked to first compare high-score prompts to low-score ones, and then elicit general ideas and learnings, and finally draft a new and better prompt. CoT further ensures the optimizer to harness collective strength from the history and identify a promising path through comparing the divergent past prompts. 

\paragraph{Flexible task prompt template:} Instead of only tuning instruction text and fixing the input position as in existing approaches such as OPRO, \ours optimizes the task prompt structure using a template that can freely and naturally move around input and instruction in the prompt. It uses placeholders for the input and any external data. For example, we instruct LLM to generate example placeholder \texttt{INSERT\_EXAMPLES\_HERE} to indicate the position of \ac{ICL} examples. In \ac{RAG} settings, we introduce a context placeholder \texttt{INSERT\_CONTEXT\_HERE} which will be replaced by the retrieved context for each question. When filled the placeholders with proper data, the task prompt clearly organized all the information to help $\text{LLM}_{\texttt{task}}$ better solve the task.



\subsection{Multi-Metric Automatic Suffix Tuning}
\label{sec:multi-metric}

Using components in Section~\ref{sec:critique} and \ref{sec:prompt-optimization}, \ours is ready to optimize a primary metric. To benefit from more teaching signals, e.g., completeness and faithfulness, 
here we extend \ours to multi-metric optimization by proposing a novel multi-metric learning extension named as \ac{AST}.


In \ac{AST}, we propose to optimize a suffix postscript $\sigma$ appended to $\bestprompt$, which has already been trained on certain metrics. $\bestprompt$ will remain fixed throughout the whole tuning process for a new metric to preserve most of its performance on existing metrics. $\bestprompt$ is extended with an additional suffix $\sigma^*\leftarrow\fcrispo(\sigma_0)$, which serves as a postscript to steer the \ac{LLM} toward the new metric and remedy any potential regression in performance on existing metrics. Specifically, we provide both the main prompt $\bestprompt$ and each suffix $\sigma_t$ in the meta-prompts while asking the LLM to critique or refine only the suffix. 
To ensure we maintain existing metrics while improving on the additional metric, we take inspirations from the balance terms of loss functions in multi-task learning \cite{he-choi-2021-stem} and compute an aggregated score across the multiple metrics. Since the score of each metric is on different scales and hard to estimate before training, we propose to use the average ranking of each metric as the ultimate basis to score prompt candidates in the meta-prompt.


\section{Main Experiments}

\subsection{Experiment Setup}
\label{sec:experiment-setup}


\paragraph{Datasets}
We select a diverse range of 4 summarization tasks including conventional document summarization tasks such as CNN daily mail~\citep{hermann2015teaching} (news headline summarization), and also conversation summarization tasks such as SAMSum~\citep{gliwa-etal-2019-samsum},
MeetingBank~\citep{hu-etal-2023-meetingbank}. In addition, we test on a medical-domain clinical note summarization task,
ACI-Bench \citep{yim2023aci}. Detailed data setup can be found in \Cref{sec:dataset setting}.
These tasks cover various lengths, domains and styles as summarized in Table~\ref{tab:data description}. We report ROUGE-1/2/L F-measure \citep{lin-2004-rouge}\footnote{We report additional metrics including AlignScore \citep{zha-etal-2023-alignscore} and BertScore \citep{zhang2019bertscore} in Appendix~\ref{appendix:full metrics}.} to measure output similarity to the references.



\paragraph{\acp{LLM} and Baselines}
We test our approach on state-of-the-art \acp{LLM} including proprietary models: Claude Instant \citep{claude_instant_model}, Claude3 Sonnet \citep{anthropic2024claude}, and open-source \acp{LLM}: Mistral 7B \cite{jiang2023mistral} and Llama3 8B \citep{llama3_model}. We use the same LLM for all the 3 \ours modules: task inference, critique-suggestion and receptive optimization, apart from the Llama3 setup. 
Specific hyper-parameters with ablations are detailed in Appendix~\ref{sec:crispo_setting}. 

Our baseline methods include manual prompts with zero/few-shot \ac{ICL}. These manual prompts are carefully tuned for each task to incorporate length constraints and task guidelines, and therefore establish a high bar of performance from manual prompt engineering (\Cref{appendix:manual prompts}). Given there are no existing automatic prompting results for text generation, we adapted \ac{OPRO}~\citep{yang2023large}, a competitive established approach, on our selected tasks. We use the same hyper-parameter setup in OPRO and \ours for fair comparison. 



\subsection{Main Results}


As shown in Table~\ref{tab:main_result}, across all the tasks and \acp{LLM}, \ours consistently improves over 0-shot manual prompt and \ac{OPRO} baselines. Overall, there are approximately 3-4 point improvements for all \acp{LLM}. Even the strong state-of-the-art Claude3 Sonnet model can still greatly benefit from \ours. The consistent improvement shows \ours is a more effective search method than existing method (OPRO) to unlock the full potential of these LLMs, and offers an alternative solution to the more labour-intensive manual prompt engineering.


Additionally, we found examples to be helpful as adding 3-shot \ac{ICL} significantly improves the performance. Owning to the versatile template in \ours, we can easily integrate examples and we show \ours 3-shot can further boost performance over \ours and achieves the best performance in most setups. 
It is also worth noticing that the vanilla \ours w/o \ac{ICL} can match or even outperform manual prompt with 3-shot in most datasets and setups, reducing latency and cost.
\subsection{Ablating Key Ingredients} Table~\ref{tab:ablations} shows the ablation results of \ours with Claude Instant on SAMSum dataset. We observed that the three key components in our approach, including flexible template, critique and step-by-step \ac{CoT} optimization, are essential for achieving optimal performance. Removing any of these components leads to a decrease in performance. Removing critique-suggestion module and CoT optimization altogether leads to a 5 point decrease, similar to OPRO performance. This indicates these two elements are essential to the success of \ours and flexible template is only effective when being added on top of these two elements.

    

\begin{table}[htbp]
\centering\small
\begin{tabular}{@{}cccccc@{}}
\toprule
\textbf{Method} & \textbf{Crit-Sugg} & \textbf{CoT} & \textbf{Template} & \textbf{avg} & \textbf{std} \\ \midrule
\ours          & \cmark                          & \cmark                        & \cmark                         & 44.4              & 1.9          \\
                & \xmark                           & \cmark                         & \cmark                          & 42.8              & 0.8          \\
                & \cmark                           & \xmark                         & \cmark                          & 43.9              & 0.3          \\
                & \cmark                           & \cmark                         & \xmark                          & 42.2              & 1.6          \\
                & \xmark                           & \xmark                         & \cmark                         & 37.4              & 3.4          \\
    OPRO            & \xmark                          & \xmark                        & \xmark                          & 38.1              & 1.3          \\ \toprule 
\textbf{Method} & \multicolumn{3}{c}{\textbf{Changing multi-aspect}}                              \\ 
\midrule
\ours & \multicolumn{3}{c}{free multi-aspects} & 44.4 & 1.9 \\
 & \multicolumn{3}{c}{no multi-aspects}                                & 41.1              & 0.9          \\
 & \multicolumn{3}{c}{pre-defined multi-aspects}                                    & 44.5              & 0.7          \\ \bottomrule
\end{tabular}
\caption{Ablation studies of \ours on the SAMSum dataset with Claude Instant. We report average and std deviation of the ROUGE-1 F results across 3 runs.}
\label{tab:ablations} 
\vspace{-3ex}
\end{table}
  


The key novelty in our proposed novel critique-suggestion strategy in \ours is that it has multi-aspect: i.e. the \ac{LLM} will generate multi-aspect comparison without enforcing predefined aspects. To understand the effect of the multi-aspect critique-suggestion, we provide two alternative baselines: 1. no multi-aspect: we ask \ac{LLM} to compare predictions and references in general with no explicit requirement for generating critique and suggestions along multiple dimensions/aspects. This is in line with the approach adopted by \citet{pryzant-etal-2023-automatic}. 2. predefined aspects: we carefully design dimensions potentially helpful for the summarization task and include verbosity, comprehensiveness, precision and style along with their definitions (Appendix~\ref{appendix:ablation_prompt}). The no multi-aspect critique-suggestion baseline performs significantly worse, lacking critical and targeted suggestions due to its tendency to be too general. The predefined multi-aspect approach is as effective as \ours but we see no significant improvement from explicit definitions of dimensions. This is because the critique \ac{LLM} in \ours is already able to identify relevant dimensions (such as completeness, verbosity etc. as in Table~\ref{tab:example}) for each iteration without explicit guidance.

\subsection{Qualitative Analysis and Human Evaluation}

\newcommand{\good}[1]{\textcolor[RGB]{0,102,0}{\textbf{#1}}}
\newcommand{\bad}[1]{\textcolor[RGB]{153,0,0}{\textbf{#1}}}



To qualitatively compare \ours outputs with the baselines, we conducted human evaluation on 20 examples from the SAMSum testset. We follow the procedure from \citet{liu-etal-2023-revisiting} where the reference summaries are split into atomic content units and annotators mark them as either present or missing in the prediction summary. In total, we collected 300 annotations (100 annotations $\times$ 3 annotators).  A final normalized recall score is computed with a length penalty, which indicates how similar the prediction summary is to the reference summary. In our experiment, we asked three annotators with postgraduate degrees to independently annotate the summaries with blinded setup. The inter-annotator agreement is ``almost perfect'' (0.8679 Fleiss kappa). We then took the majority vote, and calculated the final normalized recall score (human rating) using a de-correlated length penalty.

As shown in Table~\ref{tab:qualitative}, 
\ours achieves the highest rating according to our human evaluation. \Cref{tab:prompt_output} shows qualitative examples where prompts found by \ours better capture the style of the reference summaries in terms of length, what to focus on and what to skip. \ours outputs also look the most similar to the references, especially in terms of being as concise as the reference while covering all the key details. 

\begin{table}[!h]
\centering\small
\begin{tabular}{cccc}
\toprule
&\textbf{Manual} & \textbf{OPRO} & \textbf{\ours}  \\ \midrule
\textbf{Human Rating} & 0.58 & 0.59 & \textbf{0.63} \\ \bottomrule
\end{tabular}
\caption{
Human evaluation on sampled SAMSum test. 
}
\label{tab:qualitative}
\end{table}

\begin{table}[!h]
\centering\small
\begin{tabular}{m{8cm}}
\toprule 
\textbf{OPRO [Best Prompt]:} Generate a \bad{one to two sentence} summary within the \textlangle summary\textrangle\xspace tags that concisely describes the key details of the conversation and \bad{any conclusions} reached. INPUT\_DOC \\
\textbf{\ours [Best Prompt]:} The text below contains a discussion expressing several \good{key facts and events}. Your concise \good{1-sentence} summary should relate only the \good{2 most} important pieces of information stated,  \good{without assumptions or extra context}. INPUT\_DOC Write the summary within \textlangle summary\textrangle\xspace tags.\\\toprule
\textbf{OPRO [Example Output]:} Ralph asked Andrew if he heard a Polish joke, then told a joke about sinking a Polish battleship by putting it in water. Andrew responded that the joke was terrible and so unfunny that it made his mouth dry, requiring a sip of water. \\
\textbf{\ours [Example Output]:} Ralph tells Andrew a Polish battleship joke that Andrew finds unfunny. \\
\textbf{[Reference]:} Ralph told Andrew a joke.  \\ \bottomrule
\end{tabular}
\caption{Qualitatively comparing prompts found by OPRO and 
\ours on SAMSum. \ours is able to find a prompt to generate output more similar to the reference's concise style.}\label{tab:prompt_output}
\end{table}

\subsection{Quantitative Analysis of Prompt Diversity}

To verify that our design in \Cref{sec:method} leads to a better exploration of the solution space, we quantitatively analyze the diversity of prompts found by \ours and OPRO (same hyper-parameters, ~\Cref{sec:experiment-setup}) on the summarization datasets. We measure 4 aggregated properties on all task prompts explored by each method during optimization: length (number of words), vocabulary size (number of unique words used), and pairwise ROUGE-L/semantic similarity. For pairwise semantic similarity, we employ Sentence Transformers \cite{reimers-2019-sentence-bert} to obtain their embeddings and cosine distances.

As shown in Table~\ref{tab:prompt_diversity}, \ours prompts demonstrate larger variations in length and vocabulary while being less similar in lexicons and semantics, indicating its strength in exploring a larger space. We also provide a visualization of the prompts found by OPRO and \ours in~\Cref{sec:visualization_prompt_diversity}.

\begin{table}[h!]
  \centering\small
  \begin{tabular}{@{}ccccc@{}}
    \toprule
    \textbf{Dataset} & \textbf{Length$\uparrow$} & \textbf{Vocab$\uparrow$}  & \textbf{{\scriptsize ROUGE-L}$\downarrow$} & \textbf{Cosine$\downarrow$}   \\
    \midrule
    
    \textbf{CNN} &  & & & \\
    OPRO     & 41$\pm$6   & 36$\pm$5   &      57.5 &     0.93 \\
    \ours   & \textbf{149$\pm$24} & \textbf{96$\pm$12}  &      \textbf{50.3} &     \textbf{0.90}  \\
    \midrule

    \textbf{{\scriptsize MeetingBank}} &  & & & \\
    OPRO     & 31$\pm$5   & 28$\pm$4   &      44.9 &     0.84 \\
    \ours   & \textbf{216$\pm$41} & \textbf{135$\pm$19} &      \textbf{39.7} &     \textbf{0.80}  \\
    \midrule
  
    \textbf{SAMSum} &  & & & \\
    OPRO     & 34$\pm$6   & 30$\pm$5   &      57.0   &     0.94 \\
    \ours   & \textbf{172$\pm$22} & \textbf{112$\pm$12} &      \textbf{46.0}   &     \textbf{0.88} \\
    \midrule
 
    \textbf{ACI-Bench} &  & & & \\
    OPRO     & 58$\pm$11  & 46$\pm$8   &      62.7 &     0.95 \\
    \ours   & \textbf{247$\pm$40} & \textbf{117$\pm$13} &      \textbf{54.3} &     \textbf{0.93} \\

    \bottomrule
  \end{tabular}%
  \caption{Prompt diversity on 4 summarization datasets.}
  \label{tab:prompt_diversity}
\end{table}

\section{Extension with Multi-Metric Optimization}

\paragraph{\ac{AST} Setup \label{sec:ast_setup}}
In this experiment, we extend \ours with our proposed \ac{AST} to optimize multiple metrics simultaneously. Specifically, we take the best prompts optimized for ROUGE-1 F-measure from \ours with Claude Instant as the seed main prompt $\bestprompt$. We employ \ac{AST} to optimize AlignScore \cite{zha-etal-2023-alignscore} starting from a simple seed suffix $\sigma_0$: ``\textit{Every word of your summary must be faithful to the input/conversation}'' across all datasets. The AlignScore between the input text and the output summary is used as a signal reflecting the faithfulness.
With regard to baselines, we report the initial performance in ROUGE-1 F-measure and AlignScore of the seed main prompt w/ and w/o the seed suffix. We also provide a strong baseline to tune both the main prompt and its suffix together (full tuning) rather than only the suffix in \ac{AST}.

\paragraph{Results} The results for multi-metric optimization are presented in \Cref{tab:multi_metrics_result}. On all datasets, our \ac{AST} is able to optimize the new metric AlignScore with a negligible or zero regression on the existing metric ROUGE, meaning that \ac{AST} can reduce LLM hallucination while maintaining relevancy in the output. In particular, \ac{AST} dramatically improves AlignScore by $11.7$ points on CNN. Across tasks, \ac{AST} is the most effective approach to improve AlignScore while maintaining ROUGE. Among all methods, \ac{AST} is the only one that brings consistent improvement on AlignScore for every task, and achieves the best average overall improvement (by $4.3$). The main prompt w/ suffix seed prompt slightly improves AlignScore (by $1.2$) and the full-tuning baseline only meaningfully improves AlignScore on CNN and the overall improvement is marginal (by $0.7$). The superiority of \ac{AST} shows that it can robustly optimize multiple metrics across various domains.




\newcommand{\dG}[1]{\textcolor[RGB]{0,102,0}{#1}}
\newcommand{\dR}[1]{\textcolor[RGB]{153,0,0}{#1}}
\newcommand{\posdiff}[1]{{\small(\dG{$\uparrow$#1})}}
\newcommand{\negdiff}[1]{{\small(\dR{$\downarrow$#1})}}

\begin{table}[h!]
  \centering\small
  \resizebox{\columnwidth}{!}{%
  \begin{tabular}{@{}ccccc@{}}
    \toprule
    \textbf{}& \multicolumn{2}{c}{\textbf{Seed}} & \multicolumn{2}{c}{\textbf{\ours}: $ \fcrispo(\cdot)$} \\
    \cmidrule(lr){2-3} \cmidrule(lr){4-5}
    \textbf{} & \textbf{main} & \textbf{w/ suffix}  & \textbf{full} & \textbf{w/ AST}   \\
    \textbf{Dataset} & $\bestprompt$ & $\bestprompt + \sigma_0$  & $\fcrispo(\bestprompt + \sigma_0)$ & $\bestprompt+\fcrispo(\sigma_0)$   \\
    \midrule
    \textbf{CNN} &  & & & \\
    ROUGE-1 & \textbf{40.7} & 40.6& 40.6& 40.4\\
    AlignScore & 66.5 & 69.5\posdiff{3.0}& 69.6\posdiff{3.1}& \textbf{78.1}\posdiff{11.7}\\
    \midrule
    \textbf{ \scriptsize MeetingBank} &  & & & \\
    ROUGE-1 & 39.6 & \textbf{39.9}& 39.4& 39.7\\
    AlignScore & 43.6 & 43.7& 43.8& \textbf{44.4}\posdiff{0.9}\\
    \midrule
    \textbf{SAMSum} &  & & & \\
    ROUGE-1 & 45.5 & \textbf{45.9}& 45.8& 45.1\\
    AlignScore & 87.2 & 86.6\negdiff{0.6}& 86.6\negdiff{0.6}& \textbf{88.6}\posdiff{1.4}\\
    \midrule
    \textbf{ACI-Bench} &  & & & \\
    ROUGE-1 & 54.4 & \textbf{55.2}\posdiff{0.8}& 54.5& 54.3\\
    AlignScore & 66.7 & 69.0\posdiff{2.3}& 66.5& \textbf{70.0}\posdiff{3.4}\\
    \midrule
    \textbf{Average} &  & & & \\
    ROUGE-1 &  45.1&45.4&45.1&44.9\\
    AlignScore & 66.0& 67.2\posdiff{1.2}&66.7\posdiff{0.7}& \textbf{70.3}\posdiff{4.3}\\
    \bottomrule
  \end{tabular}%
  }
  \caption{Claude Instant multi-metric results on summarization tasks. In seed block, main and w/ suffix refers to the ROUGE-1-optimized prompt $\bestprompt$ and its concatenation with the manual suffix $\sigma_0$ respectively. In \ours block, full and w/ AST refers to the full-prompt tuning baseline and \ac{AST} in \Cref{sec:multi-metric} respectively. We show in the parentheses the absolute differences with main $p^*$ only when $>0.5$. Here we use \ours $\fcrispo(\cdot)$ to optimize an aggregation of two metrics.}
  \vspace{-3ex}
\label{tab:multi_metrics_result}
\end{table}

\section{Generalization to Other Tasks}
To confirm its generalizability, in this section, we apply \ours to extractive, abstractive and multi-choice QA tasks.

\paragraph{Datasets}
We benchmark \ours on 5 commonly used QA datasets, including 1) Wikipedia-based QA: Natural Questions~\cite{kwiatkowski-etal-2019-natural}, TriviaQA~\cite{joshi-etal-2017-triviaqa}, Squad \cite{rajpurkar-etal-2016-squad} 2) story-based abstractive reading comprehension: NarrativeQA \cite{kovcisky2018narrativeqa} and 3) medical domain multiple-choice QA: MedMCQA \cite{pal2022medmcqa} . For Natural Questions and TrivialQA, we also incorporate the \ac{RAG} setup to optimize the prompt template with inserted pre-retrieved contexts from each dataset. We retrieved the Wikipedia pages following~\citet{izacard-grave-2021-leveraging}.
For NarrativeQA, we use summaries as contexts.
For MedMCQA, we cast it to text generation by eliciting reasoning before the final answer.
Following the conventions, we report Exact Match for Natural Questions and TriviaQA, F1 for Squad, ROUGE-L for NarrativeQA, accuracy for MedMCQA. For efficiency, we only used a small fraction of the train and dev set for the experiments. The specific data settings are listed in~\Cref{sec:dataset setting}.

\paragraph{Results}
Similar to summarization tasks, we observe \ours significantly outperforms the manual prompt and OPRO baseline in various QA datasets as shown in Table~\ref{tab:qa_res}. For NarrativeQA, \ours brings massive improvement ($+10$ ROUGE-L) compared with baselines, achieving the new SOTA performance. For Natural Questions and TrivialQA, \ours has no issue incorporating the \ac{RAG} setup and achieving consistent improvement over the manual prompt and OPRO. Surprisingly, \ours even outperforms \ac{OPRO} on MedMCQA despite it is not designed for classification tasks.

\begin{table}[htb]
\centering \small
\resizebox{\columnwidth}{!}{%
\begin{tabular}{P{0.8cm}P{0.8cm}P{0.9cm}P{0.6cm}P{0.8cm}P{0.8cm}P{1.6cm}}
\toprule
                  &                & \multicolumn{2}{c}{\textbf{Manual}} & \multicolumn{3}{c}{\textbf{Automatic Prompt Engineering}} \\\cmidrule(lr){3-4}\cmidrule(lr){5-7}
\textbf{Task}           & \textbf{Claude}          & \textbf{0-shot}       & \textbf{64*}      & \textbf{OPRO}        & \textbf{\ours}        & \textbf{\ours 64*}        \\\midrule
NQ  & Instant & 34.0         & 33.4        & 8.0        &  36.5     & \textbf{37.8}             \\
& Sonnet       & 26.6         & 32.0        & 6.7        &  38.3     &  \textbf{38.7}           \\
        &         &              &             &             &             &                \\
T-QA &  Instant & 58.6         & 59.2        & 53.7       &  66.3      & \textbf{67.5}                \\

& Sonnet       & 58.4         & 65.0        & 41.8       &  70.6  & \textbf{72.1}        \\\midrule
                  &                & \textbf{0-shot}       & \textbf{5*}      & \textbf{OPRO}        & \textbf{\ours}        & \textbf{\ours 5*}        \\\midrule
Squad &  Instant & 79.5 & 82.5 & 78.5   & 87.8 & \textbf{89.4} \\
& Sonnet & 76.1 & 83.2 & 76.4 & 85.3 & \textbf{87.9}\\
        &         &              &             &             &             &                \\

NarQA &  Instant & 64.2 &     67.0     & 59.4   & 75.1  & \textbf{76.1}              \\
  &  Sonnet       &      64.0        &     66.7        &      58.6      &      \textbf{76.2}      &      75.2                \\ 
        &         &              &             &             &             &                \\
Med- &  Instant &  49.2  &    53.8      & 50.5 & 52.3 & \textbf{54.4}                  \\
MCQA &  Sonnet       &    49.8           &       54.4      & 57.7    & \textbf{57.9}   & 57.4                 \\
                 
\bottomrule
\end{tabular}}
\caption{Comparing \ours with manual prompts and competitive automatic prompt engineering baseline \ac{OPRO} on representative \ac{QA} benchmarks. We report Exact matching for NQ (Natural Questions) and T\-QA (TrivialQA). We report F1 for Squad, Rouge-L for NarQA (NarrativeQA), and accuracy for MedMCQA. $k$*: $k$-shot \ac{ICL} with example selection. Standard deviations can be found in \Cref{tab:qa_res_std}.}
\label{tab:qa_res}
\vspace{-3ex}
\end{table}

\section{Conclusion}
In this paper, we tackle the challenging problem of automatic prompt engineering for text generation. We propose \ours, a multi-aspect critique-suggestion guided optimizer augmented with enriched trajectory, \ac{CoT} and flexible template. Our experiments show multi-aspect critique-suggestion is critical for finding good task prompts. Overall, \ours achieves 3-4\% ROUGE score improvement and 4-5\% human rating increase compared to baseline methods for summarization, and significant improvement for \ac{QA}. We also show that \ours can effectively optimize multiple metrics through a novel suffix tuning extension \ac{AST}, and incorporate \ac{ICL} and \ac{RAG} with flexible prompt templates. Ablation studies confirm the effectiveness of all \ours components.  Human evaluation 
and quantitative analysis show \ours encourages more effective prompt exploration and the optimized prompts can better capture task requirements. 


\section*{Limitations}

The list of \acp{LLM} in our experiments is meant to be representative rather than exhaustive. We recognize that supervised fine-tuning can outperform prompt engineering on certain metrics. 
We also acknowledge the ongoing research on the limitations of automatic evaluation metrics for text generation. 
In addition, \ours could be costly in LLM API tokens especially with long input. 
Finally, while our experiments focus on summarization and QA, \ours can be adaptable to other text generation tasks, which we leave for future research. See \Cref{appendix: limitations} for a detailed limitations discussion.




{\small \bibliography{custom}}

\clearpage
\appendix
\onecolumn
\section*{Appendix}
\label{sec:appendix}

\newcommand\Tstrut{\rule{0pt}{2.6ex}}       
\newcommand\Bstrut{\rule[-1.2ex]{0pt}{0pt}} 
\newcommand\mcell[1]{\begin{tabular}{p{12.5cm}} \parbox{12.5cm}{\Tstrut #1\Bstrut} \end{tabular}}

\begin{table}[htb]
\centering\tiny

\begin{tabularx}{\textwidth}{c|c|X}
\toprule
\textbf{Flow}              & \textbf{Component}      & \textbf{Text} \\ \midrule
\multirow{4}{*}{\includegraphics[height=0.42\textheight]{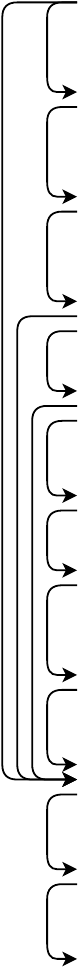}}   &\begin{tabular}[c]{@{}l@{}}  Prompt v1\end{tabular} & \mcell{Here is an input:\\\\ INSERT\_INPUT\_HERE\\\\ Write a summary within \textlangle summary\textrangle\xspace  tags for it. } \\ \cline{2-3} 
   & \begin{tabular}[c]{@{}l@{}}Output\\score: 33.3\end{tabular}         & \mcell{Tegan tells Valentia that Paul's brother sent her a friend request on social media. She is unsure how to respond to the request. Valentia advises Tegan to ignore the friend request, as she does not want Tegan to go down the same difficult road she had with Paul. Tegan agrees with Valentia's perspective.}                                                                                                                                          \\ \cline{2-3} 
  & \cellcolor{yellow!20}Critique            &\cellcolor{yellow!20}\mcell{- Number of words: The predicted summaries tended to be longer with more details while the reference summaries were shorter and more concise. \\
    - Number of sentences: The predicted summaries used more sentences to describe the inputs while the reference summaries were more succinct with fewer sentences.\\
    - Precision: Some details in the predicted summaries were not important and not mentioned in the reference summaries. \\
    - Recall: Some key details highlighted in the reference summaries were missing from the predicted summaries.
}  \\ \cline{2-3}  
                  & \cellcolor{yellow!20}Suggestion  & \cellcolor{yellow!20}\mcell{- Specifying the expected length of the summary (e.g. 1-2 sentences)\\  
- Emphasizing to only include the most important/key details\\  
- Indicating which details should be prioritized or omitted
}  \\ \cline{2-3}

&\begin{tabular}[c]{@{}l@{}}Prompt v2\end{tabular} & \mcell{For the given text, write a 1-2 sentence summary within \textlangle summary\textrangle\xspace  tags that highlights the most important details. Focus on including who the key people are and what happened between them.\\
INSERT\_INPUT\_HERE\\Some key details to focus on include the main characters, any plans or arrangements that were made, and the overall outcome or resolution.} \\ \cline{2-3} 
\cline{2-3}

                  & \begin{tabular}[c]{@{}l@{}}Output\\score: 42.1\end{tabular}         & \mcell{
Tegan tells Valentia that Paul's brother sent her a friend request on social media, though she and Paul had previously broken up. Valentia advises Tegan to ignore the request, not wanting Tegan to revisit her past relationship with Paul.}                                                                                                                                          \\ \cline{2-3} 
      & \cellcolor{yellow!20}\begin{tabular}[c]{@{}l@{}} Critique\end{tabular}              & \cellcolor{yellow!20}\mcell{- Number of words: Predicted summaries tend to be longer than reference summaries.\\
- Number of sentences: Predicted summaries use multiple sentences while references use 1-2 sentences.\\  
- Precision: Predicted summaries include more contextual details while references focus only on key people and events.\\
- Recall: Predicted summaries cover more information from the input while references only highlight the most important details.}  \\ \cline{2-3}  
                  & \cellcolor{yellow!20}\begin{tabular}[c]{@{}l@{}}Suggestion\end{tabular}    & \cellcolor{yellow!20}\mcell{- Suggest using less than 10 words where possible\\
- Emphasize concise 1-2 sentence format over detail\\
- Specify to focus only on key events/outcomes, not context}  \\ \cline{2-3}  
                  
   & \cellcolor{green!20}\begin{tabular}[c]{@{}l@{}}CoT\\ Suggestion\end{tabular}    & \cellcolor{green!20}\mcell{Comparing the high-score and low-score instructions, here are some suggestions that could improve them:\\
                \textlangle suggestion\textrangle\xspace  Specify the desired length or range for the summaries (e.g., 10 words and 1-2 sentences).\textlangle /suggestions\textrangle\xspace \\
\textlangle suggestion\textrangle\xspace  Specify to focus on key events and specify which details \textlangle /suggestion\textrangle\xspace \\
\textlangle suggestion\textrangle\xspace  Specify the output should not contain unnessary context \textlangle /suggestion\textrangle\xspace 
                }  \\ \cline{2-3}
                  & \begin{tabular}[c]{@{}l@{}}Improved\\ Prompt\end{tabular}     & \mcell{Read the dialogue provided in INSERT\_INPUT\_HERE and identify the key events between characters and outcomes. Then write a 1-2 sentence summary within \textlangle summary\textrangle\xspace  tags that concisely captures these important plot points, such as who will borrow a dress or who has an interview, while keeping within 10 words where possible. Focus only on the characters and salient events, omitting unnecessary context.}  \\ \cline{2-3} 
                  & \begin{tabular}[c]{@{}l@{}}Improved\\ Output\\
                  score: 75.6\end{tabular}     &\mcell{Tegan receives a friend request from Paul's brother and Valentia advises her to ignore it due to past issues.} \\ \midrule
                  &\begin{tabular}[c]{@{}l@{}}Reference \end{tabular}&  \mcell{Tegan has received a friend request from Paul's brother. Valentia advised her not to accept it.} \\

\bottomrule

\end{tabularx}
  \caption{ A working example for the \ours framework for SAMSum summarization with Claude3 Sonnet on one train data point. Yellow highlights the generation from the multi-aspect critique-suggestion module. Green highlights the generation from the receptive optimizer module. The score is ROUGE-1 F for the data point.}
  \label{tab:example}
\end{table}

\twocolumn

\section{A Complete Working Example}
\label{sec:working_example}
\Cref{tab:example} shows a full working example of \ours.


\section{Limitations \label{appendix: limitations}}

\paragraph{Minor prompt adaptation for different LLMs.} Different \acsp{LLM} have varying context length limits, preferred input/output formats, etc. Therefore, our approach still requires some manual adaptation to different \acsp{LLM}. However, the manual effort is significantly less compared to manually tuning task-specific prompts, because: 1) once tuned, the \texttt{crit} and \texttt{opti} prompts can be reused for different tasks, and 2) the tuning should mainly focus on formatting input/output, and adjusting the number of examples to fit the context length, which are straightforward following the \ac{LLM} documentation.

\paragraph{Evaluation metrics.} Evaluating text generation is a challenging problem in itself. For summarization, our work focuses on ROUGE scores to quantify the similarity between generated and reference texts, and AlignScore to evaluate the factuality of the generated text. We also conducted human evaluation to verify our findings. \cite{augenstein2023factuality} calls out that current factuality evaluations are not reliable. \cite{elangovan2024considers} highlights the challenges of conducting human evaluation in LLM era. However, we acknowledge that these evaluations are still limited, while designing better evaluation metrics is beyond the scope of this paper. 

\paragraph{Comparing to SOTA SFT models.}
We would like to emphasize that \ours is not designed to outperform the state-of-the-art gradient-based supervised fine-tuning (SFT) models. For some datasets, our approach still falls short compared to SOTA SFT models. Prompt tuning is a discrete optimization process with noisy directional signals on top of a limited number of prompt tokens, compared to supervised fine-tuning, which uses continuous gradient descent on much larger datasets to optimize much more parameters. Therefore, it is usually harder to match the performance of SFT.

\paragraph{Comparison between \acp{LLM}.} Our benchmark on various \acp{LLM} is designed to demonstrate that \ours is compatible with a wide range of both proprietary and open-weight (lightweight) LLMs. The list of \acp{LLM} is meant to be representative instead of exhaustive. We acknowledge the existence of more powerful \acp{LLM} from each family that may push the performance even higher, which we leave for future work.

\paragraph{Generalization beyond summarization and QA.} In our experiments, we mainly focused on summarization and question answering tasks. However, our proposed approach is general and can adapt to various text generation tasks, since the LLM-based critique-suggestion model only takes generated text and reference text as input and can spontaneously compare them along relevant dimensions. Our framework can potentially benefit classification tasks other than MedMCQA if they provide ``explanation'' or ``reasoning'' to each label.

\paragraph{Cost} Despite \ours has been optimized to use a relatively smaller number of candidates than existing methods per each step, it still requires a full evaluation of the candidates on the sampled training set of 50-200 examples, which costs significant amounts of LLM API tokens and time considering the optimization runs for 100 iterations. Especially when the inputs involve long contexts (e.g., RAG) and/or the training set is large. In our RAG settings, the optimization takes up to 2 days to finish.

\begin{table*}[htbp]
\centering\small
\begin{tabular}{rlcc}
\toprule
\textbf{Dataset}       & \textbf{Description}                                                  & \textbf{Input}      & \textbf{Output}  \\\midrule
CNN/DailyMail & News article headline generation.                            & 773   & 58  \\
MeetingBank   & City council meeting  (long conversation) summarization      & 3095 & 66  \\
SAMSum        & Messenger-like (short) conversation summarization           & 127    & 23   \\
ACI-Bench & Docter-patient (long) conversation medical note generation & 1372  & 476 \\\midrule
Natural Questions & Open-domain QA using RAG on Wikipedia & 20009.2 & 2.2 \\
TriviaQA & Open-domain QA using RAG on Wikipedia & 20016.4 & 2.8 \\
SQuAD & Reading comprehension on Wikipedia & 149.7 & 3.4 \\
NarrativeQA	 & Story reading comprehension	& 653.6 &  5.0\\
MedMCQA	& Multiple-choice QA in medical domain	& 38.0 & 100.6 \\
\bottomrule
\end{tabular}
\caption{Dataset description and the number of words in input and output. For QA datasets, the input includes both contexts and the question.}\label{tab:data description}
\end{table*}

\begin{figure*}[htbp]
    \centering
    \begin{subfigure}[b]{0.45\textwidth}
        \centering
        \includegraphics[width=\textwidth]{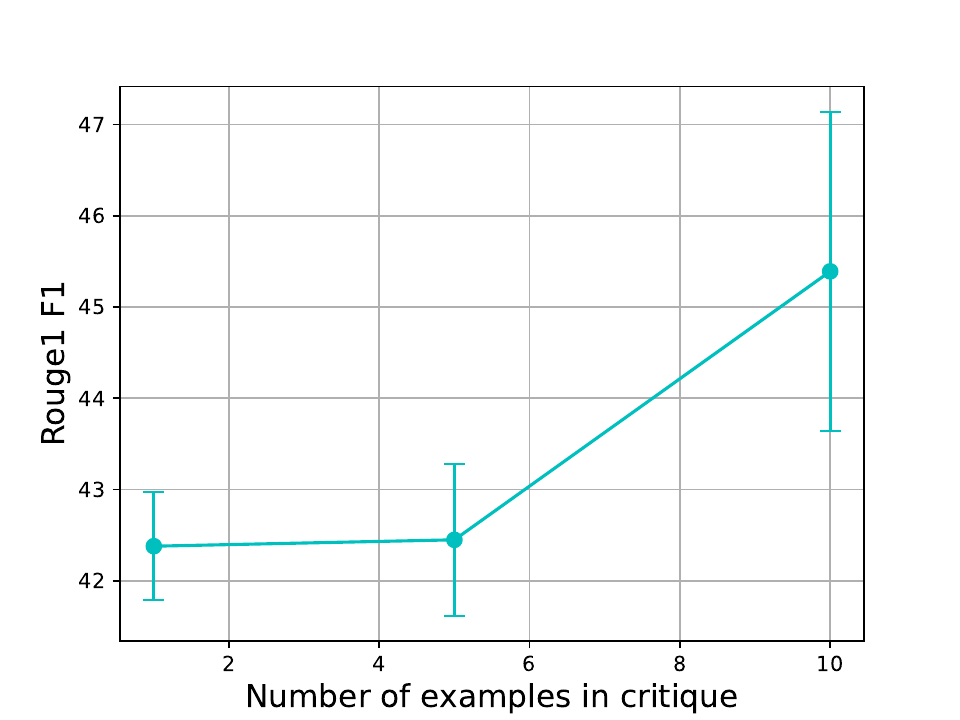}
        \caption{The effect of number of examples in the Critique-Suggestion meta-prompt}
        \label{fig:sub1}
    \end{subfigure}
    \begin{subfigure}[b]{0.45\textwidth}
        \centering
        \includegraphics[width=\textwidth]{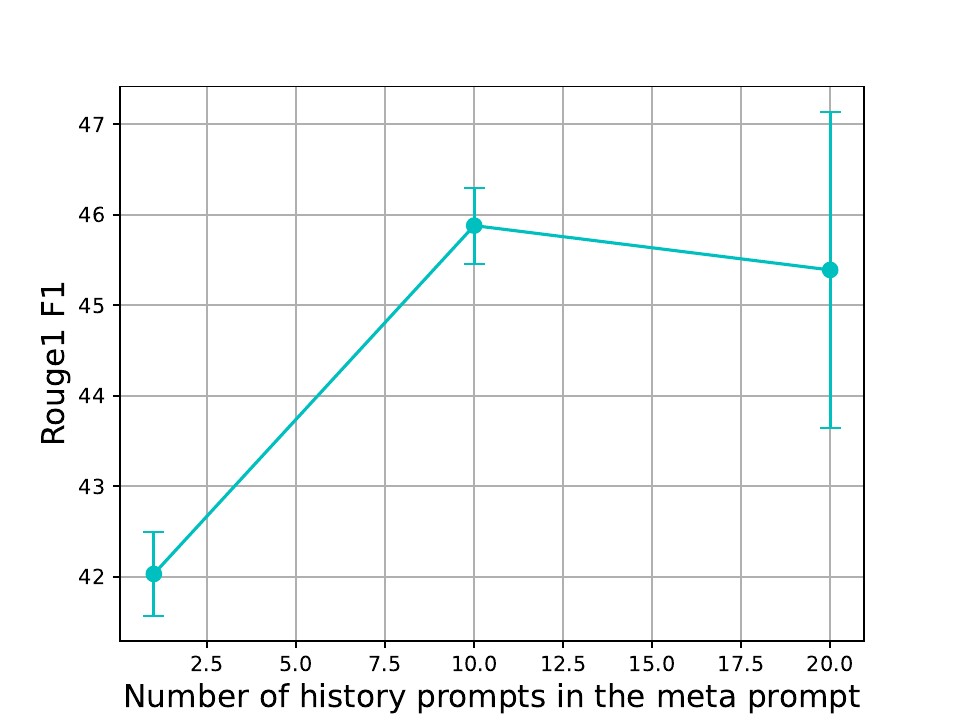}
        \caption{The effect of number of the history prompts in the receptive optimizer meta-prompt}
        \label{fig:sub2}
    \end{subfigure}
    \\
    \begin{subfigure}[b]{0.45\textwidth}
        \centering
        \includegraphics[width=\textwidth]{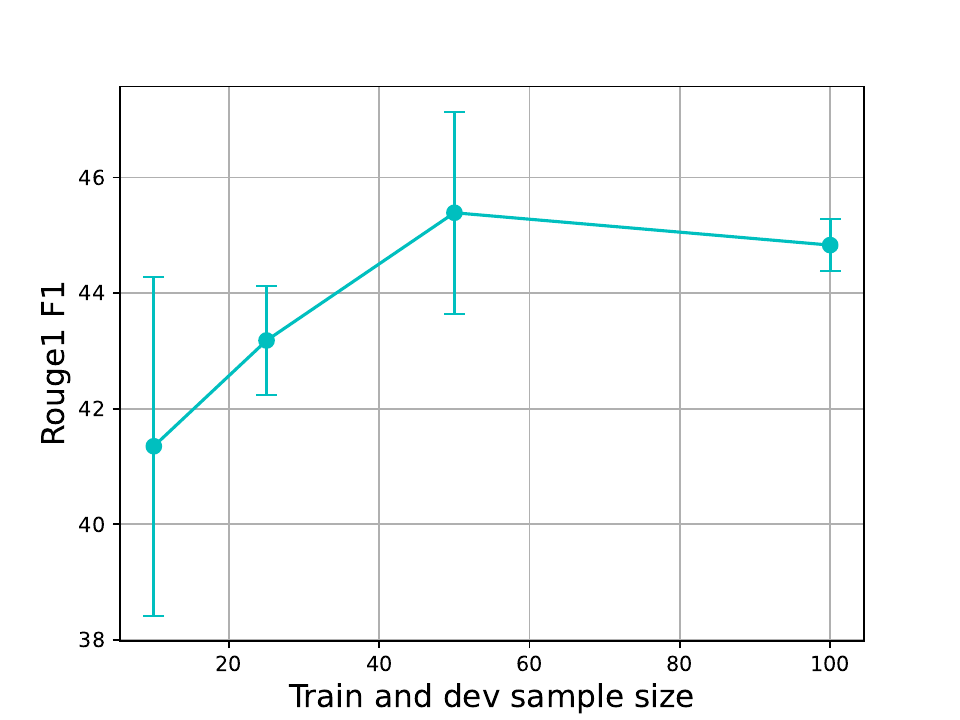}
        \caption{The effect of the train/dev sample size}
        \label{fig:sub3}
    \end{subfigure}
    \begin{subfigure}[b]{0.45\textwidth}
        \centering
        \includegraphics[width=\textwidth]{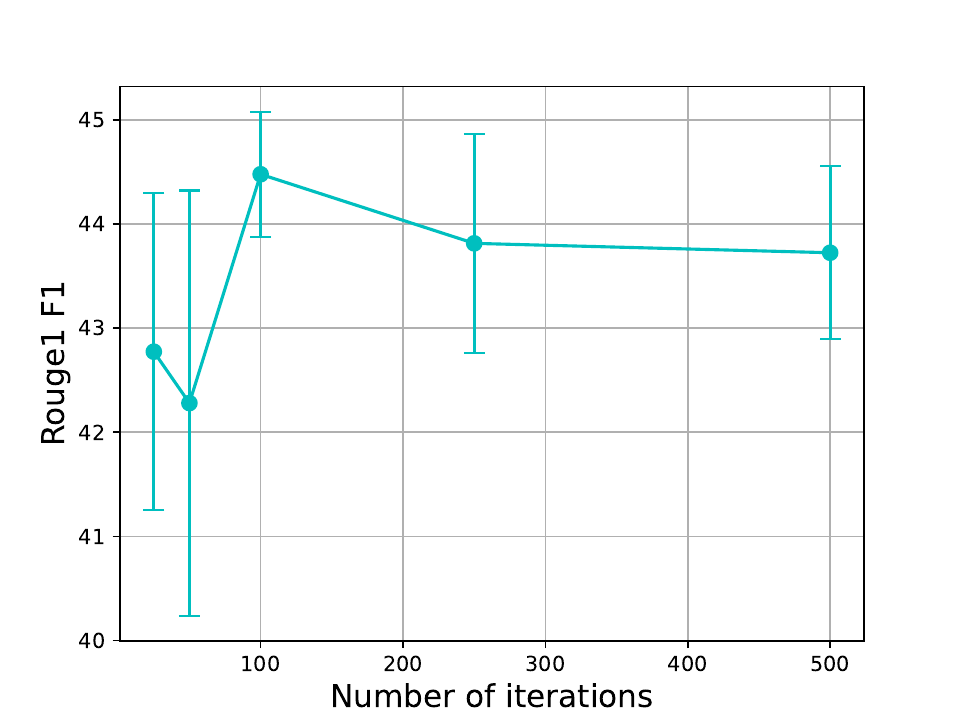}
        \caption{The effect of the number of iterations}
        \label{fig:sub4}
    \end{subfigure}
    \caption{Ablation studies with Claude Instant on SAMSum. For each setup, we report the mean ROUGE1 F1 and standard deviation across three runs. For (c), we change the random seed to select different samples across the three runs.}
    \label{fig:ablation}
\end{figure*}

\begin{figure*}[htbp]
 \centering
 \begin{subfigure}[b]{0.24\textwidth}
 \centering
 \includegraphics[width=\textwidth]{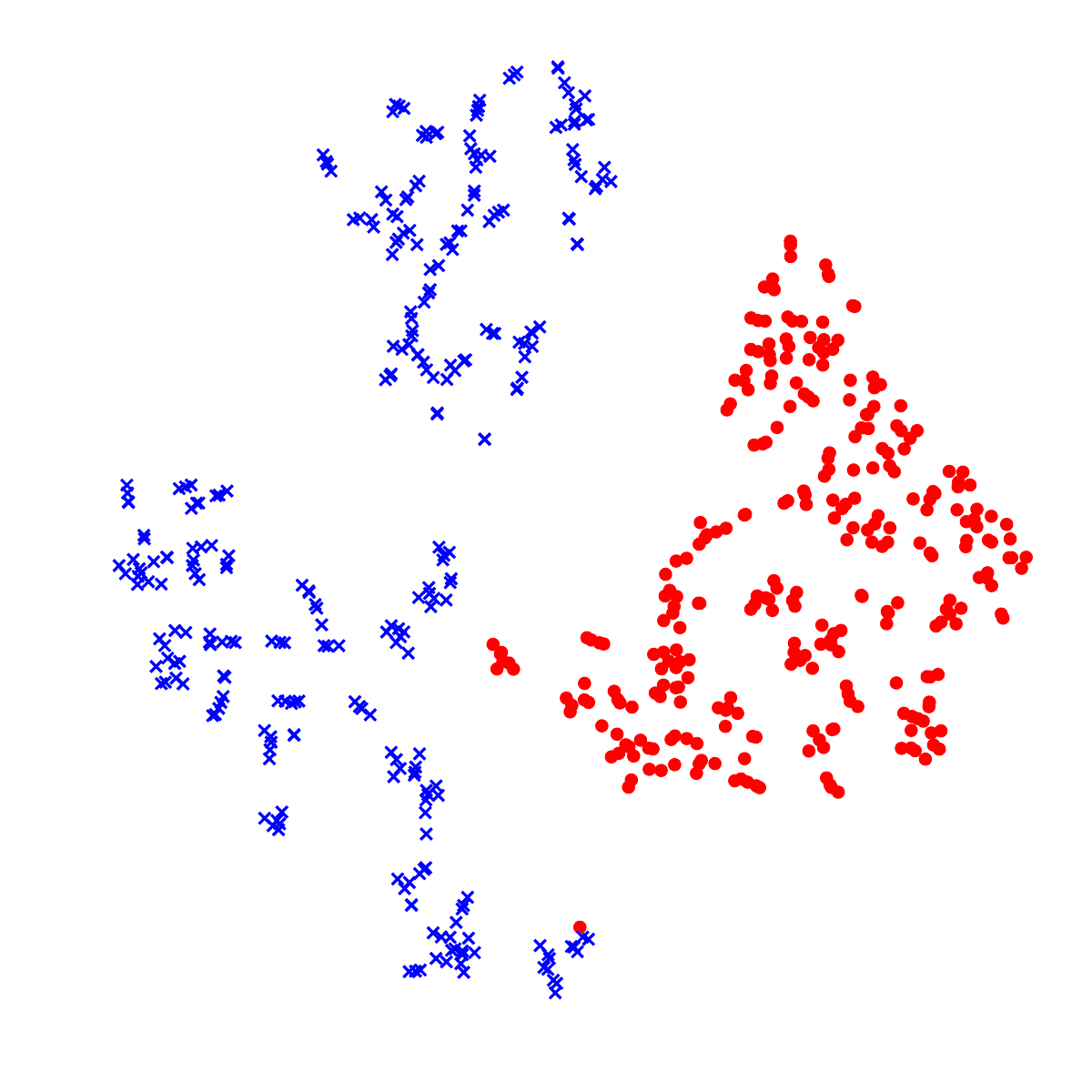}
 \caption{CNN}
 \label{fig:cnn-prompts}
\end{subfigure}
 \hfill
\begin{subfigure}[b]{0.24\textwidth}
 \centering
 \includegraphics[width=\textwidth]{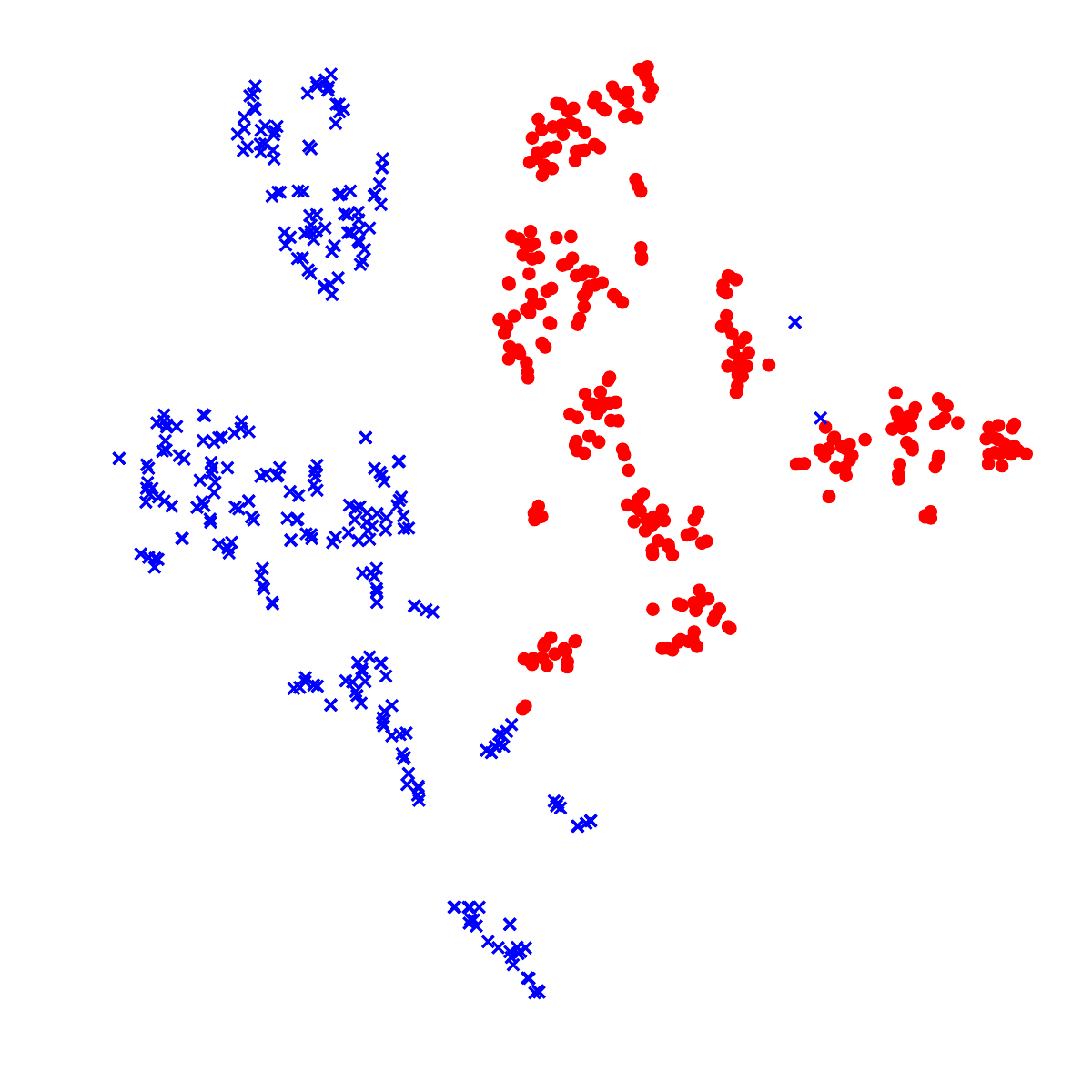}
 \caption{MeetingBank}
 \label{fig:meetingbank-prompts}
\end{subfigure}
 \hfill
\begin{subfigure}[b]{0.24\textwidth}
 \centering
 \includegraphics[width=\textwidth]{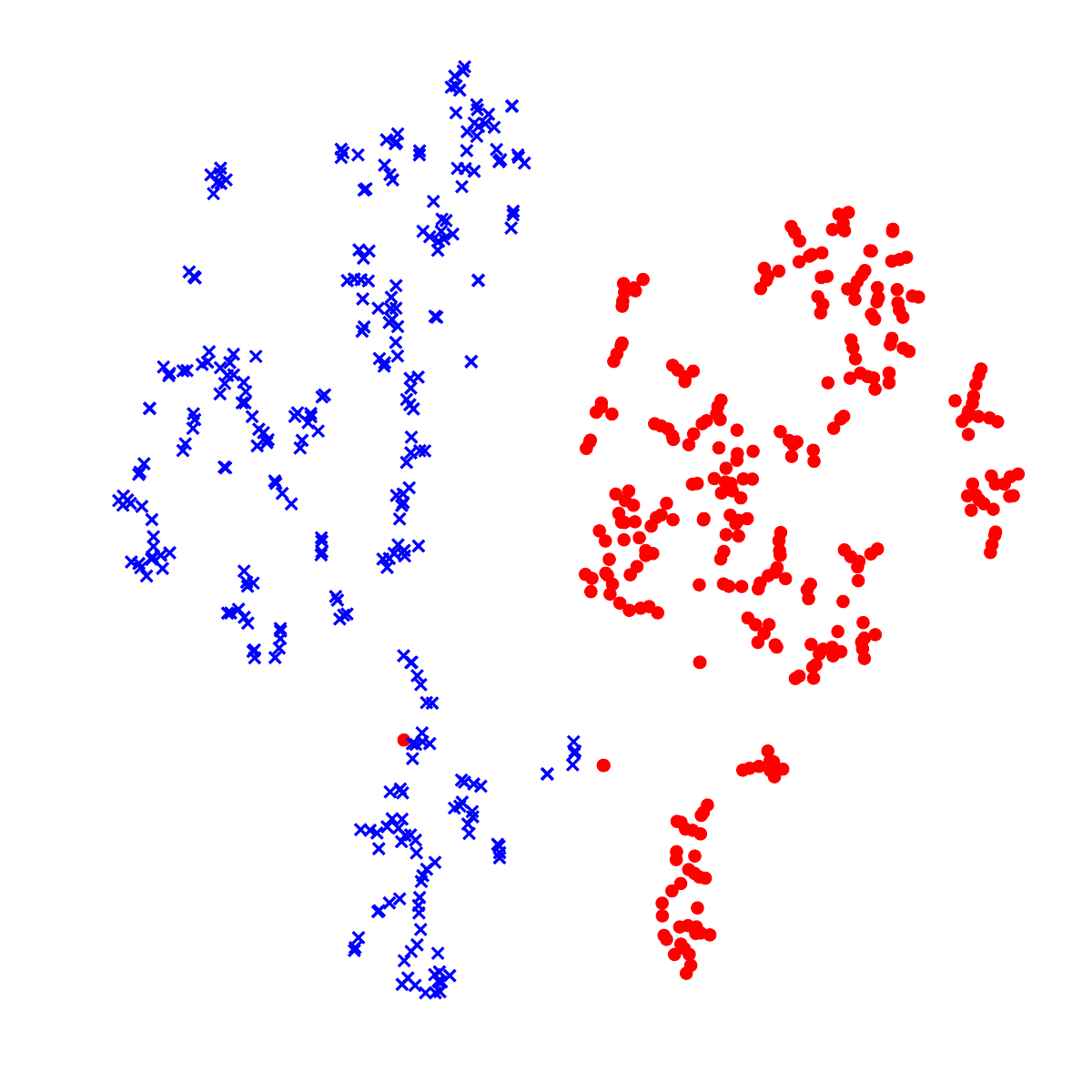}
 \caption{SAMSum}
 \label{fig:samsum-prompts}
\end{subfigure}
 \hfill
\begin{subfigure}[b]{0.24\textwidth}
 \centering
 \includegraphics[width=\textwidth]{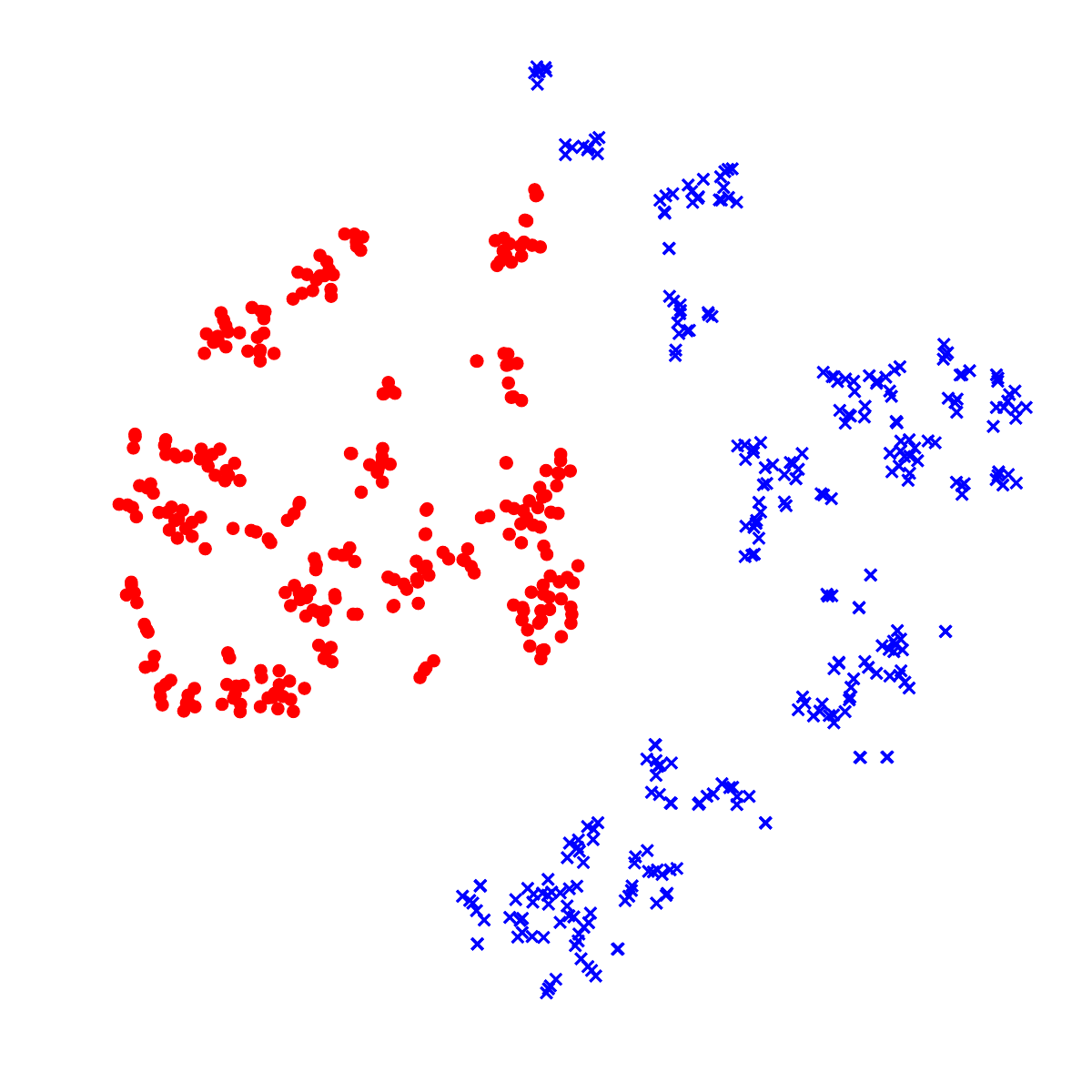}
 \caption{ACI-Bench}
 \label{fig:clinical_note_sum-prompts}
\end{subfigure}

\caption{Visualization of prompt diversity on 4 summarization datasets for \textcolor{red}{OPRO} in \textcolor{red}{red $\bullet$} and \textcolor{blue}{\ours} in \textcolor{blue}{blue $\times$}.}
\label{fig:prompt-clusters}
\end{figure*}

\section{Dataset Setting \label{sec:dataset setting}}
For CNN, SAMSum, MeetingBank, MedMCQA, Narrative QA and SQUAD, we use the HuggingFace datasets repository. For ACI-Bench, we use the data from Task B at ACL ClinicalNLP MEDIQA\-Chat shared task 2023 in the Aci\-bench dataset\footnote{\url{https://github.com/abachaa/MEDIQA-Chat-2023}}~\citep{yim2023aci}. For Natural Questions, we follow the data preparation in FiD\footnote{\url{https://github.com/facebookresearch/FiD}}~\cite{izacard-grave-2021-leveraging}.

Our experiments are conducted with sampled train and dev set. For ACI-Bench, we used the full training (67), development (20) and test set (40). For other summarization tasks, we randomly selected 500 samples from the full test set as our test set. To show the efficiency of our approach, we used a small fraction of the train and development set. For CNN, we sampled 100 training samples as our training set, and 100 development samples as our development set. For other tasks, we randomly sampled 50 training samples as our training set, and 50 development samples as our development set. 

For NQ and TQA, we randomly sample 200/200/500 examples for training/development/test set. Each example has 100 context paragraphs from Wikipedia and each paragraph has 100 words following~\citet{izacard-grave-2021-leveraging}. We use only the top 20 context paragraphs in our experiments because of the high inference cost for long text. 

For NarrativeQA and MedMCQA, we randomly sample 100/100/500 for training/development/test set respectively. For Squad, we sample 50/50/500 for training/development/test set respectively. 


\section{\ours Settings for Different LLMs}\label{sec:crispo_setting}
\textbf{Claude Settings:} In suggestion-critique meta-prompt, we pass 10 randomly selected examples for the LLM to provide critique. In optimizer meta-prompt, we use 10 history task prompts with their critiques, suggestions and scores. We add 2 input/output examples in the optimizer prompt. 

\noindent\textbf{Mistral Settings:} Mistral has a shorter context window. Therefore, we adjust the settings. We reduce the task prompt history to 1 in optimizer meta-prompt. On MeetingBank dataset, we truncate the input document to 3500 words, and do not provide the input (only use generated text and reference) in the critique-suggestion meta-prompt.

\noindent\textbf{Llama3 Settings:} The context window of Llama3 is insufficient to fit a few examples and generate meaningful critique-suggestions. Therefore, we use Claude3 Sonnet as the critique-suggestion LLM and the receptive optimizer LLM. Llama3 is used only as the task LLM.

For all experiments, we set the temperature of the meta-prompt \acp{LLM} used for the optimization to be 1.0 to encourage diversity, and we set the scorer \ac{LLM}'s temperature to be 0 which gives more stable results when the LLM is performing inference with the task prompt. We use the same LLM as meta-prompt and task prompt except for the Llama 3.

The initial prompts are generic and naive prompt. For example, for summarization, the starting prompt is ``Generate a summary for the input text''. For QA, the starting point is ``Answer the question using the context provided''. We also follow the original OPRO paper \citep{yang2023large} to sample k prompt candidates at each step for \ours. We set k = 3 in our experiments. For efficiency, we also run dev set evaluation every 5 steps rather than on every step. We end the optimization process when we reach 100 steps. 

\section{Hyper-parameter Search\label{appendix:hyperperameter}} We conducted experiments as shown in \Cref{fig:ablation} to assess the effect of different hyper-parameters.  We show the performance increases as we increase the number of examples in the critique-suggestion meta-prompt and 10 examples have a significant jump of performance compared with 1 or 5 examples in the prompt. In line with \citet{yang2023large}, we also found the history of prompts is helpful, where we see significant improvement when we increase from 1 prompt history to 10 prompt history, but no significant difference as we further increase the history to 20. As to the sample size, the performance grows at the beginning when we increase the size from 10 to 50, and plateaus moving from 50 to 100. We choose 50 sample for most of our experiments as it is a sufficiently representative sample to achieve good performance with relatively lower latency. We observed larger variations and generally lower performance when the number of iterations are below 100, but further iterations above 100 also does not further improve results as it is most likely that we already found the best prompt within 100 iterations. 
In conclusion, the most optimal combination of these hyper-parameters are: 10 examples in the critique prompt, 10 or 20 history prompts, with 50 train/dev set, with 100 iterations.  

\section{Visualization of Prompt Diversity}
\label{sec:visualization_prompt_diversity}

To verify that our design in Sec.~\ref{sec:method} leads to a better exploration of the solution space, 
we examine the distributions of prompts found by \ours and OPRO in an embedding space. We first encode their Claude Instant prompts on the 4 summarization datasets using the \texttt{all-MiniLM-L6-v2} model from Sentence Transformers \cite{reimers-2019-sentence-bert}. Then, we perform t-SNE visualization \cite{van2008visualizing} to their embeddings in a two-dimensional map.

As illustrated in \Cref{fig:prompt-clusters}, \ours produces more diverse prompts than OPRO on all of the 4 datasets. The distribution of OPRO prompts is more centralized, indicating that OPRO prompts are homogeneous in semantics, possibly the ``semantically similar paraphrases'' \cite{yang2023large}. However, \ours distribution is more divergent, with prompts spread out over a wider range. This visualization suggests that prompts tuned by \ours are semantically more dispersed and versatile, which expand the exploration beyond paraphrasing and directionless Monte-Carlo search.

\section{Experiments on Other Tasks \label{appendix:nlg_other}}

We conduct experiments on other natural language generation (NLG) tasks -- DailyDialog~\cite{li2017dailydialog} and WebNLG~\cite{gardent2017creating}. The results are shown on \cref{tab:other_nlg}.
\begin{table}[htb]
\centering\small
\begin{tabular}{ccc}
\toprule
\textbf{Method}        & \textbf{DailyDialog (R-L)} &\textbf{ WebNLG (BLEU)} \\\midrule
\textbf{Manual 0-shot} & 12.6              & 31.3          \\
\textbf{Manual 3-shot} & 17.1              & 34.8          \\
\textbf{OPRO}          & 13.3              & 33.3          \\
\textbf{CriSPO}        & 17.4              & 44.3          \\\bottomrule
\end{tabular}
\caption{Results on DailyDialog and WebNLG.}\label{tab:other_nlg}
\end{table}

\section{Multi-Aspect Critique-Suggestion Meta-Prompt}
\label{sec:critique-prompt}
\subsection{Claude for Summarization}
\begin{lstlisting}
In a summarization task, a writer is given an input text to write a summary following an instruction.

<instruction>{instruction}</instruction>
<examples>
<example>
<input>
{document}
</input>
<predicted_summary>
{predicted_summary}
</predicted_summary>
<reference_summary>
{reference_summary}
</reference_summary>
</example>
...
</examples>

Write a general and helpful critique in <critique> XML tags to improve the instruction such that the predicted summaries are as close to references as possible.

1. Come up with several dimensions to compare its predicted summaries and reference summaries, e.g., number of words, number of sentences, style, precision, recall, etc.
2. List the difference predicted summaries and references on each dimension.
3. Identify specific phrases in the instruction that could have gotten these predicted summaries different with references on each dimension.
4. Suggest specific action items that are general to all examples and helpful to improve the instruction.
\end{lstlisting}

\subsection{Mistral for Summarization}
\begin{lstlisting}
In a summarization task, a writer is given an input text to write a summary following an instruction.

INSTRUCTION: 
{instruction}

Here are a few examples using the instruction. 
EXAMPLE {id}
INPUT:
{document}
PREDICTED_SUMMARY:
{predicted_summary}
REFERENCE_SUMMARY:
{reference_summary}
...

Write a general and helpful critique to improve the instruction such that the predicted summaries are as close to references as possible.

1. Come up with several dimensions to compare its predicted summaries and reference summaries, e.g., number of words, number of sentences, style, precision, recall, etc.
2. List the difference predicted summaries and references on each dimension.
3. Identify specific phrases in the instruction that could have gotten these predicted summaries different with references on each dimension.
4. Suggest specific action items that are general to all examples and helpful to improve the instruction.
\end{lstlisting}

\subsection{Claude for RAG}
\begin{lstlisting}
In a question-answering task, question and context are provided and the answer needs to be generated.

<instruction>{instruction}</instruction>
<examples>
<example>
<question>
{question}
</question>
{context}
<generated_answer>
{generated_answer}
</generated_answer>
<gold_answer>
{gold_answer}
</gold_answer>
</example>
...
</examples>

Write a general and helpful critique in <critique> XML tags to improve the instruction such that the generated answer are the same as gold answer.

1. Come up with several dimensions to compare its generated and gold answer, e.g., number of words, style, precision, recall, etc.
2. List the difference between generated and gold answer on each dimension.
3. Identify specific phrases in the instruction that could have gotten these generated answer different with gold one on each dimension.
4. Suggest specific action items that are general to all examples and helpful to improve the instruction.
\end{lstlisting}

\section{Receptive Optimizer Meta-Prompt} 
\label{sec:meta-prompt}
\subsection{Claude for Summarization}
\begin{lstlisting}
Your task is to optimize the instruction for a summarization task, where a writer is given an input text to write its summary following your instruction.

Below are some examples:
<example>
<instruction>?</instruction>
<input>
{article}
</input>
<summary>
{summary}
</summary>
</example>
...

Below are some previous instructions with their scores and critiques.
<rated_instruction>
<instruction>{instruction}</instruction>
<score>{score}</score>
<critique>
{critique}
</critique>
</rated_instruction>
...

Generate an instruction that is different from all the instructions above, and has a higher score than all the instructions above.
It should be concise, effective, and generally applicable to all examples above.

Draft your new instruction step by step:

1. Compare high-score instructions to low-score ones, identify what suggestions could have improved them. List them in <suggestion> tags.
2. Apply the suggestions and draft a new instruction aiming for a higher score.
3. Be creative and vary the wording, paraphrase, position of INSERT_INPUT_HERE and INSERT_EXAMPLES_HERE, phrase order, grammar, sentence order and etc.
4. Write your final new instruction in <instruction> tags.
\end{lstlisting}

\subsection{Mistral for Summarization}
\begin{lstlisting}
Your task is to optimize the instruction for a summarization task, where a writer is given an input text to write its summary following your instruction.

Below are some examples:
EXAMPLE {id}
INPUT:
{article}
TARGET_SUMMARY:
{summary}
...

Below are some previous instructions with their scores and critiques.
INSTRUCTION:
{instruction}
SCORE:
{score}
CRITIQUE:
{critique}
...

Generate an instruction that is different from all the instructions above, and has a higher score than all the instructions above.
It should be concise, effective, and generally applicable to all examples above.

Draft your new instruction step by step:

1. Compare high-score instructions to low-score ones, identify what suggestions could have improved them. Write down your suggestions first.
2. Apply the suggestions and draft a new instruction aiming for a higher score.
3. Be creative and vary the wording, paraphrase, position of <INSERT_INPUT_HERE> and <INSERT_EXAMPLES_HERE>, phrase order, grammar, sentence order and etc.
4. Write your final new instruction in <instruction></instruction> tags.
5. In your final prompt, you must use <INSERT_INPUT_HERE> only once and use it in a separate line.
6. In your final prompt, you must use <INSERT_EXAMPLES_HERE> only once and use it in a separate line.
\end{lstlisting}

\subsection{Claude for RAG}
\begin{lstlisting}
Your task is to optimize the instruction for a question-answering task, where the question and context are provided.

Below are some examples:
<example>
<instruction>?</instruction>
<question>
{question}
</question>
{context}
<answer>
{answer}
</answer>
</example>
...

Below are some previous instructions with their scores and critiques.
<rated_instruction>
<instruction>{instruction}</instruction>
<score>{score}</score>
<critique>
{critique}
</critique>
</rated_instruction>
...

Generate an instruction that is different from all the instructions above, and has a higher score than all the instructions above.
It should be concise, effective, and generally applicable to all examples above.

Draft your new instruction step by step:

1. Compare high-score instructions to low-score ones, identify what suggestions could have improved them. List them in <suggestion> tags.
2. Apply the suggestions and draft a new instruction aiming for a higher score.
3. Be creative and vary the wording, paraphrase, position of "{question_placeholder}", "{context_placeholder}", phrase order, grammar, sentence order, which specific examples to give, etc.
4. Write your final new instruction in <instruction> tags.
\end{lstlisting}

\section{Manual Prompts \label{appendix:manual prompts}}
We present the manual prompts for the summarization experiments with the Claude instant model. 
INSERT\_INPUT\_HERE in each prompt indicates the position where we will insert the input text.
INSERT\_EXAMPLES\_HERE indicates the position where we will insert few-shot examples. Each example is in the format of 

\begin{lstlisting}
<examples>
<input> ...<input>
<summary> ... <summary>
</examples>
\end{lstlisting}

For the few-shot setup, we first encode inputs with BERT embeddings~\citep{devlin-etal-2019-bert}, then retrieve their most similar examples from the train set according to the cosine similarity~\citep{liu-etal-2022-makes}.

\subsection{Zero-shot CNN}
 
\begin{lstlisting}
Here is an input CNN news document:
INSERT_INPUT_HERE
Please write a headline summary between around 50 to 100 words within <summary> tags.
\end{lstlisting}

\subsection{Few-shot CNN}
 
\begin{lstlisting}
Write a headline summary between around 50 to 100 words for the CNN news document. Here are example input documents and example output summaries

INSERT_EXAMPLES_HERE

Here is an input CNN news document:

INSERT_INPUT_HERE

Please write a headline summary between around 50 to 100 words within <summary> tags.
\end{lstlisting}

\subsection{Zero-shot SAMSum}
\begin{lstlisting}
Here is an input conversation:
INSERT_INPUT_HERE
Please write a summary for the input conversation within <summary> tags. The summary should (1) be rather short with 20 to 50 words, (2) extract important pieces of information, (3) include names of interlocutors, (4) be written in the third person.
\end{lstlisting}

\subsection{Few-shot SAMSum}
\begin{lstlisting}
Write a summary within <summary> tags for the input conversation. Here are example input conversations and example output summaries

INSERT_EXAMPLES_HERE

Here is the input conversation:
INSERT_INPUT_HERE

Following the examples, please write a summary for the input conversation within <summary> tags. The summary should (1) be rather short with 20 to 50 words, (2) extract important pieces of information, (3) include names of interlocutors, (4) be written in the third person.
\end{lstlisting}

\subsection{Zero-shot MeetingBank}
\begin{lstlisting}
Here is an input conversation from city council meeting:
INSERT_INPUT_HERE
Please write a summary of the discussion with around 60 to 150 words within <summary> tags.
\end{lstlisting}

 \subsection{Few-shot MeetingBank}
\begin{lstlisting}
Write a summary for the input city council meeting. Here are example input meeting conversations and example output summaries

INSERT_EXAMPLES_HERE

Here is an input conversation from a city council meeting:

INSERT_INPUT_HERE

Following the examples, please write a summary of the discussion from the input conversation with around 60 to 150 words within <summary> tags.
\end{lstlisting}
 
\subsection{Zero-shot ACI-Bench}
\begin{lstlisting}
Here is an input conversation of a clinical visit:
INSERT_INPUT_HERE
Please write a detailed clinical note summary for the input conversation within <summary> tags.
\end{lstlisting}

\subsection{Few-shot ACI-Bench}
 
\begin{lstlisting}
Write a clinical note summary within <summary> tags for the input conversation of a clinical visit. Here are example input conversations and example output summaries

INSERT_EXAMPLES_HERE

Here is the input conversation:
INSERT_INPUT_HERE
Following the examples, please write a clinical note summary for the input conversation within <summary> tags.
\end{lstlisting}

\subsection{Manual Prompt Tuning}
While it is not possible to exhaust all prompt variations with manual prompt engineering, we experimented with several iterations of manual prompts and presented the best prompt results. Below, we show that our tuned zero-shot manual prompts (ours) significantly outperform zero-shot naive prompts ("Write a summary for the input text"), and the results from our manual prompts can be regarded as a reasonable baseline from human prompt engineering.  

\begin{table}[!h]
\centering\small
\begin{tabular}{ccccc}
    \toprule
    & \textbf{CNN} & \textbf{MBank} & \textbf{SAMSum} & \textbf{ACI-Bench}\\
    \midrule
    Naive & 34.8	&29.7	&29.9	&34.3\\
    Ours & {\bf 37.5}&	{\bf 30.7}	&{\bf 33.9}&	{\bf 43.8}\\
\bottomrule
\end{tabular}
\caption{Comparing our zero-shot manually tuned prompts (ours) with naive prompts on summarization tasks with Claude Instant. MBank= MeetingBank.}
\end{table}

\section {Best QA Prompts Found using \ours (Claude Instant)}
\subsection{Natural Questions}
\begin{lstlisting}
Consider INSERT_QUESTION_HERE and all provided INSERT_CONTEXT_HERE. Write a concise answer in <answer> tags focusing only on the single most important attribute implied across contexts. Then compare your answer to the gold below through reasoning: cite how your intended meaning matches theirs on attributes like level of precision/detail implied jointly by contexts. It is acceptable for your answer to have less context than the gold if the meaning remains clear, like using a single word versus a phrase. Explain any differences using specific examples from contexts. Answers should be as concise as possible while still encompassing implications as fully as contexts allow.
\end{lstlisting}

\subsection{TriviaQA}
\begin{lstlisting}
Read the question and contexts carefully. Extract the key detail(s) directly answering the question from the most relevant context(s). Write your response in <answer> tags matching the style and level of detail of the example gold answers. Consider using a single word, number, or short phrase if that fully answers the question precisely. Compare your answer to the examples, considering alternatives suggested in the contexts and relationships between entities. Aim for consistency with the gold answers in terms of words used, precision, and completeness of specification.

INSERT_CONTEXT_HERE

INSERT_QUESTION_HERE
\end{lstlisting}

\subsection{MedMCQA}
\begin{lstlisting}
QUESTION_PLACEHOLDER Provide your answer, and comprehensively reason through it by: referencing authoritative medical sources, accounting for all relevant context in the question, logically laying out your reasoning steps, and addressing any applicable exceptions or nuances. Your response should demonstrate a rigorous application of established medical knowledge.

Chose an option and write it in <answer> XML tags
\end{lstlisting}
\subsection{NarrativeQA}
\begin{lstlisting}
Provide a focused, concise answer in the form of a 1-3 word phrase or brief quote, enclosed in <answer> tags. Capture all key details directly relevant to fully addressing the question, while excluding extraneous background information or repetition of context details. If a short quote from the context directly and precisely answers the question in a maximally concise manner, use the quote verbatim. Otherwise, paraphrase the essential information as succinctly as possible. The goal is a clear, to-the-point response that comprehensively answers the core of the question without omitting crucial details or including unnecessary information.

CONTEXT_PLACEHOLDER

QUESTION_PLACEHOLDER
\end{lstlisting}
\subsection{Squad}
\begin{lstlisting}
INSERT_CONTEXT_HERE

INSERT_QUESTION_HERE

Your task is to answer the question as concisely as possible using only the minimum information explicitly asked for. Carefully examine the question to understand exactly what specific detail is being requested, then scan the context to extract only that precise piece of information to satisfy the question - no more and no less. Avoid including any additional context, descriptors or embellishments beyond the single term or brief phrase strictly necessary to directly answer what is asked. Refer to the examples, where "pub landlord" and "French alone is the official language" are the minimum possible responses. Do not exceed these examples in length or level of detail. Write only the clearest, most succinct answer in <answer> tags.
\end{lstlisting}

\section{Ablation Study Prompts \label{appendix:ablation_prompt}}
Pre-defined multi-aspect critique-suggestion meta-prompt:
\begin{lstlisting}
- Verbosity and length: compare the level of details and the length between prediction and reference summaries
- Comprehensiveness: compare whether the prediction covers all the information from the reference summaries
- Precision: compare whether the information from the prediction summaries are present in the reference summaries.
- Style: compare the formatting, formality, word choices, sentence structures etc.  
\end{lstlisting}

\section{Full Metrics for Summarization \label{appendix:full metrics}}
We report the average and standard deviation from 3 runs for Rouge1 (\Cref{tab:main_result_rouge1}), Rouge2 (\Cref{tab:main_result_rouge2}), RougeL (\Cref{tab:main_result_rougel}), BertScore (\Cref{tab:main_result_bertscore}) and AlignScore (\Cref{tab:main_result_alignscore}).

\begin{table*}[h!]
  \centering\small
  \begin{tabular}{@{}ccccccccc@{}}
    \toprule
    & & \multicolumn{2}{c}{\textbf{Manual}} & \multicolumn{3}{c}{\textbf{Automatic Prompt Engineering}} \\
    \cmidrule(lr){3-4} \cmidrule(lr){5-7}
     \textbf{Dataset} & \textbf{\ac{LLM}} & \textbf{0-shot} & \textbf{3-shot*}  & \textbf{OPRO} & \textbf{\ours} & \textbf{\ours 3-shot*} \\
    \midrule
    CNN & Claude Instant & 37.5 & 40.4  & 39.5 (\textpm 0.4) & 40.1 (\textpm 0.5) & 42.1 (\textpm 0.6)\\
    SOTA: 48.2& Claude3 Sonnet & 38.8 & 40.3 & 39.7 (\textpm 0.6) & 42.2 (\textpm 0.9) & 41.6 (\textpm 1.0) \\
    ~\citep{mu-lim-2022-universal} & Mistral {\scriptsize 7B} & 30.9 & 30.7 & 36.5 (\textpm 1.8) & 38.5 (\textpm 1.7) & 38.5 (\textpm 1.0) \\
    & Llama3 {\scriptsize 8B} & 37.9 & & 39.1 (\textpm 0.3)\textsuperscript{\#} & 41.5 (\textpm 0.7)\textsuperscript{\#} &\\
    \addlinespace
    MeetingBank & Claude Instant & 30.7 & 34.2 & 39.0 (\textpm 6.1) & 41.4 (\textpm 2.4) & 50.1 (\textpm 0.6)  \\
    SOTA: 70.3& Claude3 Sonnet & 31.2 & 37.5 & 41.5 (\textpm 2.2) & 47.4 (\textpm 1.7) & 58.5 (\textpm 1.3) \\
    ~\citep{hu-etal-2023-meetingbank} & Mistral {\scriptsize 7B} & 26.0 & 31.3 & 33.9 (\textpm 3.7)  &  39.1 (\textpm 4.8) & 35.2 (\textpm 0.7)  \\
    & Llama3 {\scriptsize 8B} & 31.4 & & 40.2 (\textpm 3.0)\textsuperscript{\#} & 44.7 (\textpm 0.8)\textsuperscript{\#} &\\
    \addlinespace
    SAMSum & Claude Instant & 33.9 & 37.8 & 38.1 (\textpm 1.3) & 44.4 (\textpm 1.9) & 45.8 (\textpm 0.4)  \\
    SOTA: 55.3& Claude3 Sonnet & 35.8 & 41.1 & 39.0 (\textpm 1.4) & 43.4 (\textpm 2.1) & 47.2 (\textpm 0.3)  \\
    ~\citep{wang-etal-2023-instructive} & Mistral {\scriptsize 7B} & 32.0 & 39.5 & 37.9 (\textpm 0.8)  & 37.6 (\textpm 3.4) & 40.0 (\textpm 1.0) \\
    & Llama3 {\scriptsize 8B} & 35.7 & & 39.3 (\textpm 0.6)\textsuperscript{\#} & 44.8 (\textpm 3.4)\textsuperscript{\#} &\\
    \addlinespace
    ACI-Bench & Claude Instant & 43.9 & 51.5 & 45.2 (\textpm 0.2) & 53.0 (\textpm 0.4) & 58.2 (\textpm 1.8)  \\
    SOTA: 53.5& Claude3 Sonnet & 47.3 & 59.1 & 48.8 (\textpm 1.9) & 54.0 (\textpm 1.5) &   63.1 (\textpm 0.6)\\
    ~\cite{yim2023aci} & Mistral {\scriptsize 7B} & 47.8 & 48.4 & 45.1 (\textpm 0.6)   & 50.2 (\textpm 3.0)   & 50.3 (\textpm 0.5)\\
    & Llama3 {\scriptsize 8B} & 50.5 & & 54.2 (\textpm 0.8)\textsuperscript{\#} & 56.2 (\textpm 0.4)\textsuperscript{\#} &\\

    \bottomrule
  \end{tabular}%
  \caption{Comparing \ours with manual prompts and competitive automatic prompt engineering baseline \ac{OPRO} on representative summarization benchmarks (ROUGE-1 F averaged across 3 runs). 3-shot*: 3-shot \ac{ICL} with example selection. Llama3 results marked with (\#) is using Claude-3 as the optimizer, due to limited context window of Llama3. }
  \label{tab:main_result_rouge1}
\end{table*}

\begin{table*}[h!]
  \centering\small
  \begin{tabular}{@{}ccccccccc@{}}
    \toprule
    & & \multicolumn{2}{c}{\textbf{Manual}} & \multicolumn{3}{c}{\textbf{Automatic Prompt Engineering}} \\
    \cmidrule(lr){3-4} \cmidrule(lr){5-7}
     \textbf{Dataset} & \textbf{\ac{LLM}}& \textbf{0-shot} & \textbf{3-shot*}  & \textbf{OPRO} & \textbf{\ours} & \textbf{\ours 3-shot*} \\
    \midrule
  CNN& Claude Instant& 12.5 &14.8 &14.3 (\textpm 0.3)&15.7 (\textpm 0.9)&17.0 (\textpm 0.2) \\
& Claude3 Sonnet& 14.4 &15.4 &15.1 (\textpm 0.2)&17.3 (\textpm 1.5)&16.3 (\textpm 0.5) \\
& Mistral {\scriptsize 7B} & 11.0 &10.6 &14.4 (\textpm 0.8)&14.3 (\textpm 0.6)&14.3 (\textpm 0.1) \\
& Llama3 {\scriptsize 8B} & 14.4 &&15.2 (\textpm 0.4)\textsuperscript{\#}&16.3 (\textpm 0.9)\textsuperscript{\#}& \\
\addlinespace
MeetingBank& Claude Instant& 11.6 &17.3 &20.3 (\textpm 6.9)&23.7 (\textpm 4.7)&35.4 (\textpm 0.5) \\
& Claude3 Sonnet& 14.2 &22.0 &21.8 (\textpm 2.8)&32.5 (\textpm 2.2)&46.5 (\textpm 1.8) \\
& Mistral {\scriptsize 7B} & 11.5 &14.8 &15.4 (\textpm 2.5)&19.5 (\textpm 6.7)&16.7 (\textpm 0.9) \\
& Llama3 {\scriptsize 8B} & 14.6 &&22.3 (\textpm 2.7)\textsuperscript{\#}&27.6 (\textpm 0.4)\textsuperscript{\#}& \\
\addlinespace
SAMSum& Claude Instant& 11.7 &14.3 &13.4 (\textpm 0.9)&16.9 (\textpm 2.2)&18.7 (\textpm 0.8) \\
& Claude3 Sonnet& 12.7 &16.6 &14.7 (\textpm 0.1)&17.1 (\textpm 1.0)&20.8 (\textpm 0.3) \\
& Mistral {\scriptsize 7B} & 10.2 &14.1 &13.6 (\textpm 1.4)&12.4 (\textpm 1.5)&14.2 (\textpm 1.0) \\
& Llama3 {\scriptsize 8B} & 12.3 &&14.7 (\textpm 0.4)\textsuperscript{\#}&18.8 (\textpm 3.8)\textsuperscript{\#}& \\
\addlinespace
ACI-Bench& Claude Instant& 16.9 &23.6 &16.3 (\textpm 0.4)&19.7 (\textpm 0.6)&26.7 (\textpm 2.3) \\
& Claude3 Sonnet& 20.3 &30.1 &20.1 (\textpm 1.4)&21.4 (\textpm 0.8)&32.5 (\textpm 0.9) \\
& Mistral {\scriptsize 7B} & 17.7 &19.2 &17.0 (\textpm 0.1)&18.2 (\textpm 1.7)&18.7 (\textpm 0.7) \\
& Llama3 {\scriptsize 8B} & 19.8 &&22.0 (\textpm 0.2)\textsuperscript{\#}&22.8 (\textpm 0.2)\textsuperscript{\#}& \\

    \bottomrule
  \end{tabular}%
  \caption{Comparing \ours with manual prompts and competitive automatic prompt engineering baseline \ac{OPRO} on representative summarization benchmarks (ROUGE-2 F averaged across 3 runs). 3-shot*: 3-shot \ac{ICL} with example selection. Llama3 results marked with (\#) is using Claude3 Sonnet as the optimizer, due to limited context window of Llama3. }
  \label{tab:main_result_rouge2}
\end{table*}

\begin{table*}[h!]
  \centering\small
  \begin{tabular}{@{}ccccccccc@{}}
    \toprule
    & & \multicolumn{2}{c}{\textbf{Manual}} & \multicolumn{3}{c}{\textbf{Automatic Prompt Engineering}} \\
    \cmidrule(lr){3-4} \cmidrule(lr){5-7}
     \textbf{Dataset} & \textbf{\ac{LLM}}& \textbf{0-shot} & \textbf{3-shot*}  & \textbf{OPRO} & \textbf{\ours} & \textbf{\ours 3-shot*} \\
    \midrule
  CNN& Claude Instant& 22.6 &24.8 &24.5 (\textpm 0.5)&26.1 (\textpm 0.4)&27.4 (\textpm 0.5) \\
& Claude3 Sonnet& 24.0 &25.2 &25.1 (\textpm 0.5)&27.9 (\textpm 0.9)&27.1 (\textpm 0.6) \\
& Mistral {\scriptsize 7B} & 20.4 &20.1 &23.0 (\textpm 1.5)&23.9 (\textpm 1.3)&24.1 (\textpm 0.7) \\
& Llama3 {\scriptsize 8B} & 23.8 &&24.6 (\textpm 0.4)\textsuperscript{\#}&26.5 (\textpm 0.5)\textsuperscript{\#}& \\
\addlinespace
MeetingBank& Claude Instant& 20.5 &25.5 &29.7 (\textpm 7.4)&33.1 (\textpm 4.5)&44.4 (\textpm 0.2) \\
& Claude3 Sonnet& 22.3 &29.5 &32.0 (\textpm 2.8)&40.9 (\textpm 2.0)&54.1 (\textpm 1.6) \\
& Mistral {\scriptsize 7B} & 18.5 &22.7 &24.2 (\textpm 3.4)&29.3 (\textpm 6.5)&26.1 (\textpm 1.0) \\
& Llama3 {\scriptsize 8B} & 22.6 &&31.5 (\textpm 3.3)\textsuperscript{\#}&36.8 (\textpm 0.7)\textsuperscript{\#}& \\
\addlinespace
SAMSum& Claude Instant& 25.6 &28.8 &28.7 (\textpm 1.2)&34.3 (\textpm 2.0)&36.2 (\textpm 0.2) \\
& Claude3 Sonnet& 27.0 &31.3 &30.1 (\textpm 1.1)&34.3 (\textpm 2.3)&38.2 (\textpm 0.5) \\
& Mistral {\scriptsize 7B} & 24.1 &30.3 &29.0 (\textpm 0.7)&28.4 (\textpm 2.9)&30.8 (\textpm 1.3) \\
& Llama3 {\scriptsize 8B} & 27.1 &&30.0 (\textpm 0.5)\textsuperscript{\#}&35.4 (\textpm 3.4)\textsuperscript{\#}& \\
\addlinespace
ACI-Bench& Claude Instant& 26.1 &33.5 &25.5 (\textpm 1.0)&26.8 (\textpm 1.4)&35.3 (\textpm 2.3) \\
& Claude3 Sonnet& 29.3 &38.6 &29.5 (\textpm 1.1)&30.3 (\textpm 0.4)&41.0 (\textpm 0.6) \\
& Mistral {\scriptsize 7B} & 25.4 &28.1 &25.2 (\textpm 0.1)&25.6 (\textpm 1.9)&26.2 (\textpm 0.4) \\
& Llama3 {\scriptsize 8B} & 27.7 &&29.3 (\textpm 0.6)\textsuperscript{\#}&29.9 (\textpm 0.5)\textsuperscript{\#}& \\

    \bottomrule
  \end{tabular}%
  \caption{Comparing \ours with manual prompts and competitive automatic prompt engineering baseline \ac{OPRO} on representative summarization benchmarks (ROUGE-L F averaged across 3 runs). 3-shot*: 3-shot \ac{ICL} with example selection. Llama3 results marked with (\#) is using Claude3 Sonnet as the optimizer, due to limited context window of Llama3. }
  \label{tab:main_result_rougel}
\end{table*}

\begin{table*}[h!]
  \centering\small
  \begin{tabular}{@{}ccccccccc@{}}
    \toprule
    & & \multicolumn{2}{c}{\textbf{Manual}} & \multicolumn{3}{c}{\textbf{Automatic Prompt Engineering}} \\
    \cmidrule(lr){3-4} \cmidrule(lr){5-7}
     \textbf{Dataset} & \textbf{\ac{LLM}}& \textbf{0-shot} & \textbf{3-shot*}  & \textbf{OPRO} & \textbf{\ours} & \textbf{\ours 3-shot*} \\
    \midrule
  CNN& Claude Instant& 87.0 &87.6 &87.5 (\textpm 0.1)&87.2 (\textpm 0.4)&87.7 (\textpm 0.3) \\
& Claude3 Sonnet& 87.4 &87.7 &87.5 (\textpm 0.0)&87.8 (\textpm 0.0)&87.8 (\textpm 0.3) \\
& Mistral {\scriptsize 7B} & 85.6 &85.8 &87.0 (\textpm 0.1)&87.3 (\textpm 0.2)&87.3 (\textpm 0.1) \\
& Llama3 {\scriptsize 8B} & 87.2 &&87.4 (\textpm 0.1)\textsuperscript{\#}&87.6 (\textpm 0.1)\textsuperscript{\#}& \\
\addlinespace
MeetingBank& Claude Instant& 85.0 &86.0 &86.7 (\textpm 1.2)&86.8 (\textpm 0.3)&89.2 (\textpm 0.1) \\
& Claude3 Sonnet& 85.4 &86.9 &87.1 (\textpm 0.4)&88.1 (\textpm 0.3)&90.8 (\textpm 0.3) \\
& Mistral {\scriptsize 7B} & 84.3 &85.3 &85.8 (\textpm 0.7)&86.2 (\textpm 0.3)&85.9 (\textpm 0.2) \\
& Llama3 {\scriptsize 8B} & 85.4 &&86.7 (\textpm 0.6)\textsuperscript{\#}&87.7 (\textpm 0.2)\textsuperscript{\#}& \\
\addlinespace
SAMSum& Claude Instant& 89.2 &89.8 &89.8 (\textpm 0.2)&90.4 (\textpm 0.4)&90.7 (\textpm 0.5) \\
& Claude3 Sonnet& 89.5 &90.3 &89.8 (\textpm 0.4)&90.6 (\textpm 0.7)&91.3 (\textpm 0.1) \\
& Mistral {\scriptsize 7B} & 88.3 &90.0 &89.8 (\textpm 0.2)&89.5 (\textpm 0.6)&90.1 (\textpm 0.2) \\
& Llama3 {\scriptsize 8B} & 88.7 &&89.9 (\textpm 0.1)\textsuperscript{\#}&90.7 (\textpm 0.5)\textsuperscript{\#}& \\
\addlinespace
ACI-Bench& Claude Instant& 85.5 &88.1 &85.1 (\textpm 0.3)&85.8 (\textpm 0.7)&88.1 (\textpm 0.5) \\
& Claude3 Sonnet& 85.7 &89.1 &85.7 (\textpm 0.5)&86.1 (\textpm 0.3)&90.0 (\textpm 0.3) \\
& Mistral {\scriptsize 7B} & 85.3 &86.4 &84.9 (\textpm 0.1)&85.5 (\textpm 0.8)&85.8 (\textpm 0.2) \\
& Llama3 {\scriptsize 8B} & 85.1 &&86.1 (\textpm 0.2)\textsuperscript{\#}&86.6 (\textpm 0.4)\textsuperscript{\#}& \\

    \bottomrule
  \end{tabular}%
  \caption{Comparing \ours with manual prompts and competitive automatic prompt engineering baseline \ac{OPRO} on representative summarization benchmarks (BERTScore F averaged across 3 runs). 3-shot*: 3-shot \ac{ICL} with example selection. Llama3 results marked with (\#) is using Claude3 Sonnet as the optimizer, due to limited context window of Llama3.}
  \label{tab:main_result_bertscore}
\end{table*}

\begin{table*}[h!]
  \centering\small
  \begin{tabular}{@{}ccccccccc@{}}
    \toprule
    & & \multicolumn{2}{c}{\textbf{Manual}} & \multicolumn{3}{c}{\textbf{Automatic Prompt Engineering}} \\
    \cmidrule(lr){3-4} \cmidrule(lr){5-7}
     \textbf{Dataset} & \textbf{\ac{LLM}}& \textbf{0-shot} & \textbf{3-shot*}  & \textbf{OPRO} & \textbf{\ours} & \textbf{\ours 3-shot*} \\
    \midrule
 CNN& Claude Instant& 76.1 &83.1 &85.5 (\textpm 1.3)&73.9 (\textpm 12.6)&77.8 (\textpm 7.1) \\
(Reference: 78.7)& Claude3 Sonnet& 84.5 &86.0 &84.6 (\textpm 1.3)&84.5 (\textpm 4.9)&83.9 (\textpm 5.5) \\
& Mistral {\scriptsize 7B} & 84.9 &85.2 &84.5 (\textpm 5.9)&84.4 (\textpm 1.3)&86.4 (\textpm 0.5) \\
 & Llama3 {\scriptsize 8B} & 83.7 &&85.4 (\textpm 0.9)\textsuperscript{\#}&86.1 (\textpm 1.2)\textsuperscript{\#}& \\
\addlinespace
MeetingBank& Claude Instant& 72.5 &70.8 &59.6 (\textpm 12.8)&61.9 (\textpm 6.2)&64.0 (\textpm 2.1) \\
(Reference: 51.4)& Claude3 Sonnet& 71.9 &70.8 &57.5 (\textpm 3.3)&49.9 (\textpm 16.8)&70.5 (\textpm 2.1) \\
 & Mistral {\scriptsize 7B} & 76.5 &72.1 &76.5 (\textpm 6.5)&76.5 (\textpm 2.1)&76.6 (\textpm 0.6) \\
 & Llama3 {\scriptsize 8B} & 72.2 &&71.9 (\textpm 14.9)\textsuperscript{\#}&63.7 (\textpm 1.3)\textsuperscript{\#}& \\
\addlinespace
SAMSum& Claude Instant& 85.7 &86.6 &84.5 (\textpm 0.9)&85.3 (\textpm 3.6)&83.9 (\textpm 1.2) \\
(Reference: 79.9) & Claude3 Sonnet& 87.9 &87.2 &89.5 (\textpm 0.5)&87.0 (\textpm 1.3)&84.4 (\textpm 1.5) \\
& Mistral {\scriptsize 7B} & 87.6 &86.8 &88.4 (\textpm 0.8)&87.7 (\textpm 0.7)&87.4 (\textpm 1.3) \\
& Llama3 {\scriptsize 8B} & 88.8 &&88.9 (\textpm 0.7)\textsuperscript{\#}&87.7 (\textpm 2.5)\textsuperscript{\#}& \\
\addlinespace
ACI-Bench& Claude Instant& 66.7 &66.3 &62.3 (\textpm 2.0)&63.3 (\textpm 3.3)&65.6 (\textpm 1.1) \\
(Reference: 61.4) & Claude3 Sonnet& 70.2 &67.4 &69.8 (\textpm 7.8)&65.0 (\textpm 3.3)&63.8 (\textpm 0.6) \\
& Mistral {\scriptsize 7B} & 68.0 &69.0 &65.6 (\textpm 0.4)&67.8 (\textpm 2.3)&67.2 (\textpm 1.6) \\
& Llama3 {\scriptsize 8B} & 72.5 &&59.4 (\textpm 1.8)\textsuperscript{\#}&62.3 (\textpm 1.2)\textsuperscript{\#}& \\

    \bottomrule
  \end{tabular}%
  \caption{Comparing \ours with manual prompts and competitive automatic prompt engineering baseline \ac{OPRO} on representative summarization benchmarks (\textbf{AlignScore} averaged across 3 runs). 3-shot*: 3-shot \ac{ICL} with example selection. Llama3 results marked with (\#) is using Claude3 Sonnet as the optimizer, due to limited context window of Llama3. Although AlignScore is among the SoTA metrics for the factual consistency, it is far from perfect. It tends to favor verbatim copies and simple paraphrases of the source document over highly abstractive summaries. As an evidence, the ground truth reference summary from the dataset gets lower AlignScore than LLM generated summary. }
  \label{tab:main_result_alignscore}
\end{table*}

\section{Standard Deviation for QA}

\begin{table*}[tb]
\centering
\begin{tabular}{ccccccc}
\toprule
                  &                & \multicolumn{2}{c}{\textbf{Manual}} & \multicolumn{3}{c}{\textbf{Automatic Prompt Engineering}} \\\cmidrule(lr){3-4}\cmidrule(lr){5-7}
\textbf{Dataset}           & \textbf{Claude}          & \textbf{0-shot}       & \textbf{64-shot}      & \textbf{OPRO}        & \textbf{\ours}        & \textbf{\ours 64-shot}        \\\midrule
Natural Question (Exact Match)  & Instant & 34.0         & 33.4        & 8.0 (\textpm 6.6)         &  36.5 (\textpm 2.2)       & \textbf{37.8} (\textpm 1.1)              \\
SOTA: 60.4~\cite{izacard2023atlas}  & Sonnet       & 26.6         & 32.0        & 6.7 (\textpm 5.9)         &  38.3 (\textpm 1.6)       &  \textbf{38.7} (\textpm 3.9)            \\
        &         &              &             &             &             &                \\
TriviaQA (Exact Match) &  Instant & 58.6         & 59.2        & 53.7 (\textpm 3.3)        &  66.3 (\textpm 1.1)       & \textbf{67.5} (\textpm 1.0)                 \\
SOTA: 86.1~\cite{touvron2023llama} & Sonnet       & 58.4         & 65.0        & 41.8 (\textpm 23.9)        &  70.6 (\textpm 0.2)       & \textbf{72.1} (\textpm 0.3)                 \\\midrule
                  &                & \textbf{0-shot}       & \textbf{5-shot}      & \textbf{OPRO}        & \textbf{\ours}        & \textbf{\ours 5-shot}        \\\midrule
Squad (F1)&  Instant & 79.5 & 82.5 & 78.5  (\textpm 4.1) & 87.8 (\textpm 0.5) & \textbf{89.4} (\textpm 0.2) \\
SOTA: 95.8 \citep{li-etal-2020-dice}& Sonnet & 76.1 & 83.2 & 76.4  (\textpm 7.4) & 85.3 (\textpm 3.8) & \textbf{87.9} (\textpm 2.5) \\
        &         &              &             &             &             &                \\

NarrativeQA (Rouge-L) &  Instant & 64.2 &     67.0     & 59.4 (\textpm 13.2)    & 75.1 (\textpm 0.4)  & \textbf{76.1} (\textpm 0.5)                   \\
  SOTA: 59.87 \cite{nishida-etal-2019-multi}          &  Sonnet       &      64.0        &     66.7        &      58.6 (\textpm {\color{white}0}9.9)       &      \textbf{76.2} (\textpm 1.6)       &      75.2 (\textpm 1.0)                \\ 
        &         &              &             &             &             &                \\
MedMCQA (Accuracy) &  Instant &  49.2  &    53.8      & 50.5 (\textpm 0.9)    & 52.3 (\textpm 2.9)  & \textbf{54.4} (\textpm 2.1)                   \\
  SOTA: 73.7 \cite{nori2023capabilities}          &  Sonnet       &    49.8           &       54.4      & 57.7 (\textpm 2.1)    & \textbf{57.9} (\textpm 0.9)  & 57.4 (\textpm 0.3)                   \\
                 
\bottomrule
\end{tabular}
\caption{Comparing \ours with manual prompts and competitive automatic prompt engineering baseline \ac{OPRO} on representative question answering benchmarks. *-shot: few-shot \ac{ICL} with example selection. }
\label{tab:qa_res_std}
\end{table*}

\Cref{tab:qa_res_std} shows the full results for QA (Question Answering) datasets with standard deviation reported over three runs.

\end{document}